\documentclass{article}

\usepackage{microtype}
\usepackage{graphicx}
\usepackage{subfigure}
\usepackage{booktabs} 
\usepackage{float}

\usepackage{hyperref}
\usepackage{multirow}
\usepackage{array} 
\usepackage[export]{adjustbox}

\usepackage{caption}

\usepackage[accepted]{icml2024}

\usepackage{amsmath}
\usepackage{amssymb}
\usepackage{mathtools}
\usepackage{amsthm}

\usepackage[capitalize,noabbrev]{cleveref}

\theoremstyle{plain}

\theoremstyle{definition}

\theoremstyle{remark}

\usepackage[textsize=tiny]{todonotes}

\icmltitlerunning{An Interactive Agent Foundation Model}

\begin{document}

\twocolumn[{

\icmltitle{An Interactive Agent Foundation Model}



\vspace{-4mm}
\icmlsetsymbol{equal}{*}
\icmlsetsymbol{lead}{$\blacktriangleright$}
\icmlsetsymbol{equal advisor}{$\dagger$}
\icmlsetsymbol{intern}{$\S$}
\begin{icmlauthorlist}
\icmlauthor{Zane Durante}{equal,stanford,microsoft,intern}, 
\icmlauthor{Bidipta Sarkar}{equal,stanford,microsoft,intern}, 
\icmlauthor{Ran Gong}{equal,microsoft,ucla,intern}, 
\icmlauthor{Rohan Taori}{stanford,microsoft,intern}, 
\icmlauthor{Yusuke Noda}{microsoft}, \\
\icmlauthor{Paul Tang}{stanford},  
\icmlauthor{Ehsan Adeli}{stanford}, 
\icmlauthor{Shrinidhi Kowshika Lakshmikanth}{stanford},  
\icmlauthor{Kevin Schulman}{stanford},  
\icmlauthor{Arnold Milstein}{stanford},  
\icmlauthor{Demetri Terzopoulos}{ucla}, 
\icmlauthor{Ade Famoti}{microsoft}, 
\icmlauthor{Noboru Kuno}{microsoft}, 
\icmlauthor{Ashley Llorens}{microsoft}, 
\icmlauthor{Hoi Vo}{microsoft,equal advisor}, \\ 
\icmlauthor{Katsu Ikeuchi}{microsoft,equal advisor}, 
\icmlauthor{Li Fei-Fei}{stanford,equal advisor}, 
\icmlauthor{Jianfeng Gao}{microsoft,equal advisor}, 
\icmlauthor{Naoki Wake}{equal,microsoft,lead}, 
\icmlauthor{Qiuyuan Huang}{equal,microsoft,lead}
\end{icmlauthorlist}

\icmlaffiliation{stanford}{Stanford University}
\icmlaffiliation{ucla}{University of California, Los Angeles}
\icmlaffiliation{microsoft}{Microsoft Research, Redmond}


\icmlkeywords{Machine Learning, ICML}

\vskip 0.15in

\begin{center}
    \captionsetup{type=figure}
    \includegraphics[width=0.98\linewidth]{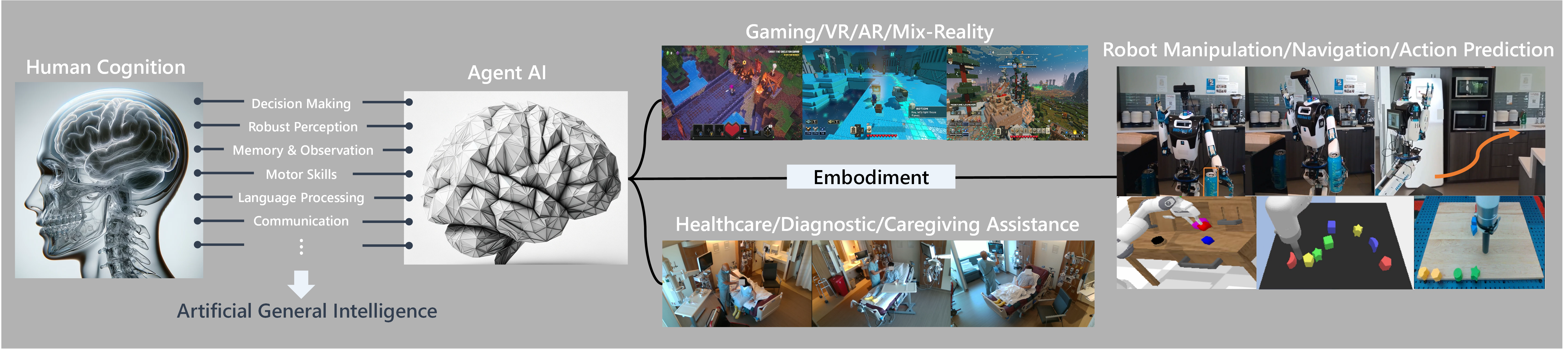}
    \vspace{-3mm}
    \captionof{figure}{Overview of an Agent AI system that can perceive and act in different domains and applications.
     Agent AI is emerging as a promising avenue toward Artificial General Intelligence (AGI).  Our model represents an initial step in the development of a model that is highly capable of human-level reasoning across many tasks and levels of granularity.}
    \label{fig:agenttask}
    \vspace{3mm}
\end{center}
}]



\let\svthefootnote\thefootnote


\begin{abstract}
The development of artificial intelligence systems is transitioning from creating static, task-specific models to dynamic, agent-based systems capable of performing well in a wide range of applications.  We propose an \textbf{Interactive Agent Foundation Model} that uses a novel multi-task agent training paradigm for training AI agents across a wide range of domains, datasets, and tasks. Our training paradigm unifies diverse pre-training strategies, including visual masked auto-encoders, language modeling, and next-action prediction, enabling a versatile and adaptable AI framework. We demonstrate the performance of our framework across three separate domains---Robotics, Gaming AI, and Healthcare. Our model demonstrates its ability to generate meaningful and contextually relevant outputs in each area. The strength of our approach lies in its generality, leveraging a variety of data sources such as robotics sequences, gameplay data, large-scale video datasets, and textual information for effective multimodal and multi-task learning. Our approach provides a promising avenue for developing generalist, action-taking, multimodal systems. 
\end{abstract}

\let\thefootnote\relax\footnotetext{$^{*}$Equal Contribution. $^{\blacktriangleright}$Project Lead. $^{\dagger}$ Equal Advisor. \\ $^{\S}$ Work done while interning or researching part-time at Microsoft Research, Redmond.
$^{1}$Stanford University; $^{2}$Microsoft Research, Redmond; $^{3}$University of California, Los Angeles.}

\section{Introduction}
\label{sec:intro}

The development of AI systems that can not only gather useful sensory information, but also interact with their environments in meaningful ways has been a long-time goal for AI researchers.  One key advantage of developing generalist AI systems is that of training a single neural model across many tasks and data modalities, an approach which is highly scalable via data, compute, and model parameters \cite{reed2022generalist}.  With recent significant advances surrounding general-purpose foundation models \cite{bommasani2021opportunities}, the AI community has a new set of tools for developing generalist, action-taking AI systems en route to artificial general intelligence.  Despite their impressive results across various AI benchmarks, large foundation models frequently hallucinate the presence of objects and actions in scenes and infer factually incorrect information \cite{rawte2023survey,peng2023check}.  We posit that one of the key reasons why these foundation models hallucinate is due to their lack of grounding in the environments in which they are trained (e.g., large-scale internet data instead of physical or virtual environments). Furthermore, the dominant approach for building multimodal systems is to leverage frozen pre-trained foundation models for each modality 
and to train smaller layers that allow for cross-modal information passing \cite{alayrac2022flamingo,li2022blip,li2023blip,dai2023instructblip,liu2023llava}. Since the visual- and language-specific submodules are not tuned during multimodal training, any hallucination errors in the submodules will likely be present in the resulting multimodal system. Additionally, lack of cross-modal pre-training could make grounding information across modalities challenging.

Towards such a generalist model that is grounded and pre-trained within physical or virtual environments, we propose a unified pre-training framework for handling text, visual data, and actions as input.  We treat each input type as separate tokens and pre-train our model to predict masked tokens across all three modalities.  Our approach uses pre-trained language models and pre-trained visual-language models to effectively initialize our model with pre-trained submodules, which we jointly train in our unified framework. We call our approach and resulting model an \textbf{Interactive Agent Foundation Model}, due to its ability to \textit{interact} with humans and its environment, as well as its 
visual-language understanding ability as shown in Figure~\ref{fig:agenttask}.   


\let\thefootnote\svthefootnote
\setcounter{footnote}{0}


In this paper, we show that a 277M parameter model\footnote{We are currently developing an even larger model.} that is jointly pre-trained across 13.4 M video frames from several distinct domains and data sources can effectively engage in interactive multi-modal settings using text, video, images, dialogue, captioning, visual question answering, and embodied actions within four disparate virtual environments. In order to effectively evaluate the broad range of capabilities and generalization abilities of our model, we show results across distinct domains: (1) Robotics, (2) Gaming AI, and (3) Healthcare.  Despite using domain-specific visual inputs, text descriptions, and action-spaces, our model is effectively able to generalize across all three domains.  To facilitate research in this discipline, we plan to release our code and models publicly. 

\section{Related Work}
\label{sec:related_works}

\subsection{Foundation Models} 

A large number of works have sought to develop general-purpose foundation models based on large-scale pre-training on broad-scale internet data from a variety of sources \cite{bommasani2021opportunities}. Within the field of Natural Language Processing, this generally consists of larger proprietary LLMs \cite{wang2022self} such as the GPT-series \cite{brown2020language,min2022rethinking}, or smaller open-source models such as the LLaMA series \cite{touvron2023llama}, or instruction-tuned variants such as Alpaca \cite{alpaca} and Vicuna \cite{zheng2023judging}.  Within the field of computer vision, strategies such as masked auto-encoders \cite{he2021masked} and contrastive learning \cite{radford2021learning} are two popular methods for self-supervised learning.


\subsection{Multimodal Understanding}



Recently, many multimodal models have been developed that seek to learn a relatively small number of parameters to connect large pre-trained visual encoders and language model decoders (that are generally frozen) with representative models including Flamingo \cite{alayrac2022flamingo}, the BLIP-series \cite{li2022blip,li2023blip,dai2023instructblip}, and LLaVA \cite{liu2023llava}. These models are generally trained using the standard language modeling cross-entropy loss on large-scale internet data consisting of visual-text pairs, using a source of data similar to that used to train contrastive dual encoder models \cite{radford2021learning,bain2021frozen,sun2023eva}. Unlike most previous work, we explore training models to predict visual tokens and action tokens in addition to language tokens and explicitly train our model for agentic tasks.


\subsection{Agent-Based AI}

\begin{figure*}[t]
    \centering
    \includegraphics[width=0.95\linewidth]{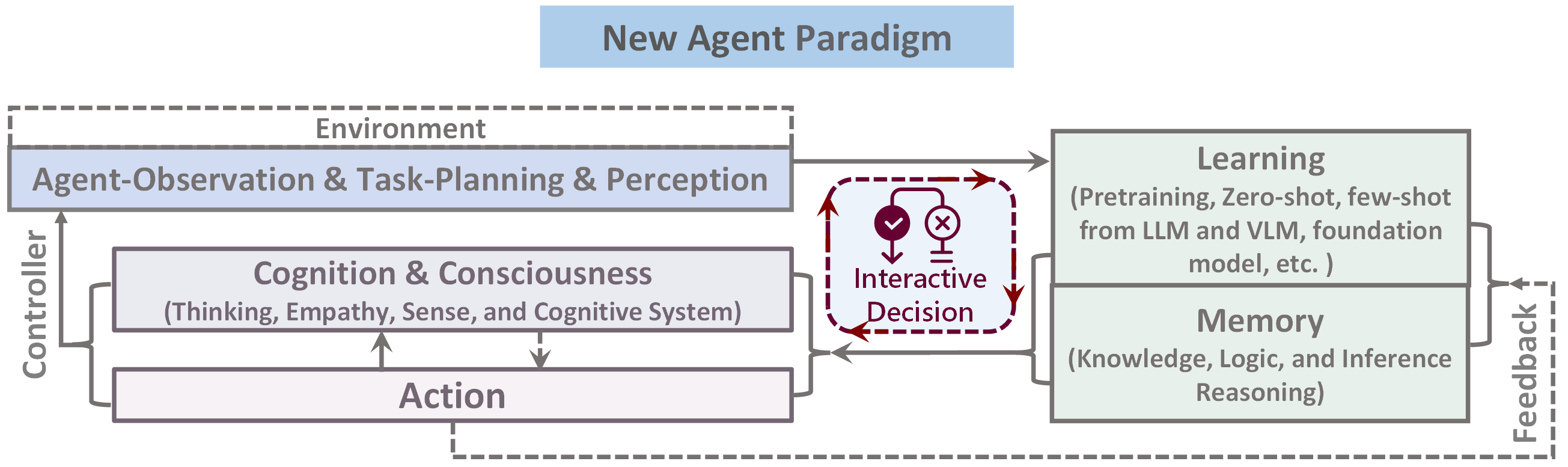}
    \vspace{-3mm}
    \caption{We propose an Agent AI paradigm for supporting interactive multi-modal generalist agent systems. There are 5 main modules as shown: (1) Agent in Environment and Perception with task-planning and observation, (2) Agent learning, (3) Memory, (4)
Action, and (5) Cognition and Consciousness (we use ``consciousness'' to imply a degree of awareness of an agent's state and surroundings). A key difference between our approach and some previous interactive strategies is that, after training, the agent's action will directly impact task planning, as the agent does not need to receive feedback from the environment to plan its next actions. }
    \label{fig:agentpara}
\end{figure*}

Agent-based AI is distinguished from traditional AI by its need to generate dynamic behaviors that are grounded in an understanding of environmental contexts. Recent research has focused on employing advanced large foundation models to create Agent-based AI systems, as shown in \cite{durante2024agent}. In the field of robotics, for instance, recent studies have highlighted the potential of LLM/VLMs in enhancing multimodal interactions between robots, environments, and humans. This applies to both manipulation \cite{jiang2022vima,brohan2022rt,pantazopoulos-etal-2023-multitask,brohan2023rt,li2023vision,ahn2022can,shah2023mutex,li2023mastering, wake2023gpt,gong2023lemma} and navigation \cite{gadre2023cows,dorbala2023can,cai2023bridging,shah2023lm,zhou2023navgpt,dorbala2022clip,liang2023mo,huang2023visual}. Additionally, significant advances in reinforcement learning have improved agent policy training on top of VLM/LLMs. Key advancements have been made in areas such as reward design \cite{yu2023language, katara2023gen2sim, ma2023eureka}, efficient data collection \cite{kumar2023words, du2023video}, and the management of long-horizon steps \cite{xu2023creative, sun2023prompt, li2023interactive, parakh2023human, wake_chatGPT}.  Similarly to robotics, gaming agents require an understanding of visual scenes and textual instructions/feedback \cite{puig2023habitat,li2021igibson,srivastava2022behavior,gong2023mindagent}. Agent-AI in the context of healthcare has focused on the text-based interaction between humans by utilizing the capabilities of LLM/VLMs. Representative applications include diagnostic assistance \cite{lee2023benefits, li2023llavamed}, knowledge retrieval \cite{peng2023check, guu2020retrieval}, and remote monitoring \cite{amjad2023review}.

\section{Agent Paradigm}
\label{sec:paradigm}

Recent advancements in AI technology have been remarkable, enabling a reasonable understanding of linguistic and visual information acquired in open-world environments. At this pivotal historical juncture, public interest in embodied agent technology is shifting from research confined to simulations and controlled environments to practical applications in highly uncertain environments. For example, consider a scenario where a robot, upon being unboxed, can instantly start communicating with non-expert humans and swiftly adapt to performing household tasks in the home environment. In this section, we define a new paradigm for embodied agents to position our proposed Interactive Agent Foundation Model within the context of this new paradigm.

We define the embodied agent paradigm  as \textit{``any intelligent agent capable of autonomously taking suitable and seamless action based on sensory input, whether in the physical world or in a virtual or mixed-reality environment representing the physical world''} (Figure~\ref{fig:agentpara}). Importantly, an embodied agent is conceptualized as a member of a \textbf{collaborative system}, where it communicates with humans with its vision-language capabilities and employs a vast set of actions based on the humans' needs. In this manner, embodied agents are expected to mitigate cumbersome tasks in virtual reality and the physical world.

\begin{figure*}[t]
    \centering
    \includegraphics[width=0.90\linewidth]{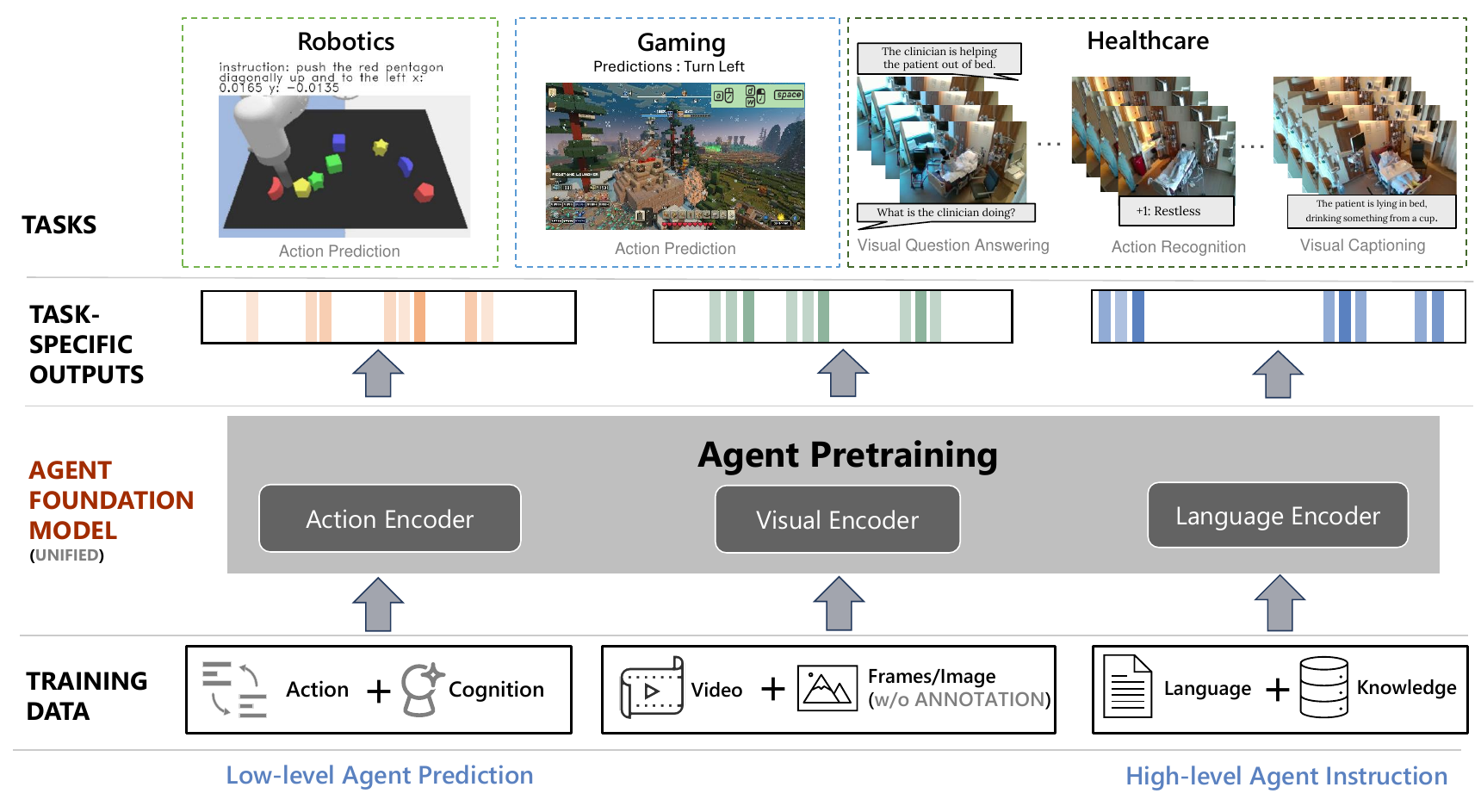}
    \vspace{-3mm}
    \caption{Overview of our Interactive Agent framework. 
    Our foundation model is designed to process multi-modal information that conveys various levels of abstraction. This approach facilitates a comprehensive understanding of the context and environment, thus ensuring that actions are coherent. By training on a variety of task domains and applications, we develop a versatile foundation model that can be fine-tuned for executing optimal actions in a variety of contexts, paving the way towards generally intelligent agents. 
    }
    \label{fig:agentfoandation}
\end{figure*}

We believe such a system of embodied agents requires at least three key components:
\begin{enumerate}
\item \textbf{Perception} that is multi-sensory with fine granularity. Like humans, multi-sensory perception is crucial for agents to understand their environment, such as gaming environments, to accomplish various tasks. In particular, visual perception is useful for agents that can parse the visual world (e.g., images, videos, gameplay). 
\item \textbf{Planning} for navigation and manipulation. Planning is important for long-range tasks, such as navigating in a robotics environment and conducting sophisticated tasks. Meanwhile, planning should be grounded on good perception and interaction abilities to ensure plans can be realized in an environment.
\item \textbf{Interaction} with humans and environments. Many tasks require multiple rounds of interactions between AI and humans or the environment. Enabling fluent interactions between them would improve the effectiveness and efficiency of completing tasks for AI.
\end{enumerate}

In light of these principles, our proposed \textbf{Interactive Agent Foundation Model} represents preliminary research that focuses on these critical aspects, aiming to develop an embodied agent that functions as a practical assistance system.  For an overview of our goals for developing an embodied agent, see Figure \ref{fig:agentpara}.

Achieving an embodied agent is not easy, especially considering the complex dynamics of systems with multi-modal observations in the physical world. Despite the advancement of recent LLM/VLMs, many challenges must be addressed, including but not limited to: 1) unstructured environments, where current visual inputs affect both high-level and low-level actions of the embodied agent given the same goal instruction; 2) open sets of objects, which require the agent's decision-making module to use common sense knowledge that is hard to encode manually; 3) natural language interactions, which require the agent to understand and operate on more than just template-based commands, but also a context of goals, constraints, and partial plans expressed in everyday language. To enable a more comprehensive approach to these complex challenges, the inclusion of researchers and practitioners from a broader range of fields is critical.



\section{Agent Foundation Model} 
\label{sec:model}
 
Our proposed framework is shown in Figure~\ref{fig:agentfoandation}. 
  By synergistically combining visual perception with linguistic understanding, our models offer the potential to endow robots with a more intuitive understanding of their surroundings and better contextual reasoning.  Our current work focuses on developing a joint image and video encoder and aligning this joint encoder to existing foundation models.  This has several notable benefits: firstly, it allows for the use of both action, image, and video with language datasets for pre-training. Secondly, it increases the capabilities of the model across a variety of downstream tasks (e.g., video understanding, temporal reasoning, action prediction, interaction with human feedback, etc.). Finally, by using a joint encoder, we can reduce the overall model size (instead of using two separate encoders), which can be useful for edge deployments or in limited computing scenarios such as robotics, gaming, and interactive healthcare tasks.

\subsection{Model Architecture}
\label{sec:architecture}

\begin{figure}
    \centering
    \includegraphics[width=\columnwidth]{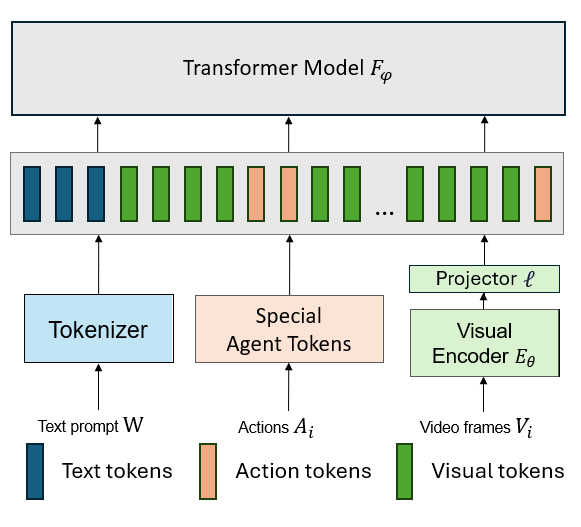}
    \vspace{-10mm}
    \caption{Our Unified Tokenization Framework. We propose a general pre-training strategy for predicting input tokens.  For text tokens, we use the standard language modeling task with next token prediction.  For actions, we expand the vocabulary of the language model to include special ``agent'' tokens that represent each of the actions available to the language model. Finally, we incorporate visual tokens into our framework by training a visual encoder to predict masked visual tokens.}
    \label{fig:token_stream}
\end{figure}

To effectively initialize our model to handle text, visual, and agent tokens as input, we initialize our architecture with two pre-trained submodules.  First, we use CLIP ViT-B16 from \cite{radford2021learning} to initialize our visual encoder, denoted $E_{\theta}$, and initialize our action and language model, $F_\phi$, from OPT-125M \cite{zhang2022opt}.  We encode each frame in a video $V_i$ as visual features $Z_i = E_\theta(V_i)$.  We enable cross-modal information sharing by training an additional linear layer $\ell$ that transforms the embeddings of our visual encoder $E_\theta$ into the token embedding space of our transformer model $F_\phi$.  Thus, given a text prompt $W$ and a single video frame $V_i$, we can obtain $\hat{A}$, a text token or action token prediction via $\hat{A} = F_\phi(W, \ell(E_\theta(V_i)))$. To incorporate prior time steps into our model, we also include the previous actions and visual frames as input during pre-training.  For a given time step $t$, we predict $\hat{A}_t$ as
\begin{multline}
    \hat{A}_t = F_\phi(W, \ell(E_\theta(V_1)), A_1,  \ell(E_\theta(V_2)), A_2, \\ \dots,  \ell(E_\theta(V_{t-1})), A_{t-1}, \ell((E_\theta(V_t))).
\end{multline}  
In practice, due to memory constraints, we only handle the previous $M$ actions and frames, and update the previous $V_i$ and $A_i$ as a sliding window. In order to  more effectively train our visual encoder to predict masked visual tokens, we use sinusoidal positional embeddings, as in \cite{he2021masked} instead of the positional embeddings of CLIP.  Since we are using relatively small checkpoints, we are able to jointly train our entire model during pre-training, unlike previous visual-language models that largely rely upon frozen submodules and seek to learn an adaptation network for cross-modal alignment  \cite{alayrac2022flamingo,li2022blip,liu2023llava}. We show our general process for formatting our input tokens in Figure~\ref{fig:token_stream}, and describe our pre-training strategy in Section~\ref{sec:pretraining-strategy}.  For additional details, see Appendix~\ref{sec:appendix_architecture}.

\subsection{Pre-Training Strategy}
\label{sec:pretraining-strategy}

We pre-train our model on a wide range of robotics and gaming tasks, with each input sample containing text instructions, videos, and action tokens. We notate each sample as a sequence $S = (W, V_1, A_1, V_2, A_2, \dots, V_T, A_T)$, where $W$ is the sequence of tokens corresponding to the text instruction, $V_i$ is the sequence of image patches corresponding to frame $i$, and $A_i$ is the sequence of action tokens corresponding to the frame $i$ of a video sequence of $T$ frames. We denote $w_j$ as the tokens of the text prompt $W$, and denote the parameters of our model as $\theta$. For each sample, there are three components to the loss function: language modeling, masked image auto-encoding, and action modeling.

The language modeling loss is a standard causal language modeling loss to minimize the negative log likelihood of each token in the instruction conditioned on prior tokens. The language modeling loss for a particular sample $S$ is
\begin{equation}
    L_{lang}(S) = -\sum_{j=1}^{|W|} \log p_{\theta}(w_j | w_{<j}).
\end{equation}

The masked image autoencoding loss is generated by randomly masking 75\% of the image patches and calculating the mean-squared error between the reconstructed image and original image in pixel space for the masked image patches. The masked auto-encoder loss for a particular sample, $S$ is: 
\begin{equation}
    L_{mae}(S) = \sum_{t=1}^{T} ||\textsc{u}(V_{t}) - \textsc{u}(D_{\theta}(E_{\theta}(\textsc{m}(V_t))))||_2^2,
\end{equation}
where $\textsc{m}$ randomly masks 75\% of the image patches, $\textsc{u}$ only selects the previously masked out features, and $E_{\theta}$ and $D_{\theta}$ are the encoder and decoder for the vision module, respectively.

Finally, the action modeling loss minimizes the negative log-likelihood of each action token conditioned on all prior information, including all text tokens, prior visual tokens, and prior action tokens. The action modeling loss for a particular sample $S$ is: \begin{equation}
    L_{act}(S) = -\sum_{t=1}^{T} \sum_{i=1}^{|A_t|} \log p_{\theta}((a_{t})_i | W, V_{\leq t}, A_{\leq t}, (a_{t})_{<i}).
\end{equation}

The full loss function for each sample combines the above components: 
\begin{equation}
    L(S) = \frac{L_{lang}(S) + L_{mae}(S) + L_{act}(S)}{|W| + \sum_{t=0}^T (|V_t| + |A_t|)}.
\end{equation}

On robotics data, we only use $T=4$ frames of video as input since the tasks are Markovian and therefore do not require long histories to accurately predict the next action. Our gaming data samples use $T=9$ frames of video as input since an observation history is necessary for the partially-observable gaming tasks.

\section{Tasks}
\label{sec:tasks}

We believe that a foundational model, trained in visual, language, and agent capabilities, leads to a powerful and general-purpose tool that significantly impacts a variety of interactive tasks. 
To evaluate the effectiveness of our approach, we applied the model to three major agent-AI scenarios, encompassing representative downstream tasks: 1) Robotics: human-machine manipulation in the physical world; 2) Gaming: human-machine embodiment in virtual reality; 3) Healthcare: augmented human-machine interaction in traditional multimodal tasks. For these tasks, the pre-trained model was fine-tuned with specific datasets. As a result, the model demonstrated reasonable and competitive performance in terms of action prediction, visual understanding, natural language-driven human-machine interactions, gaming, and hospital scene understanding. We outline the task definitions and specific datasets used below.

\subsection{Robotics Tasks}

For the robotics scenario, we tested the model on language-guided manipulation tasks. To this end, we selected two distinct robotics manipulation datasets: Language-Table \cite{lynch2023interactive} and CALVIN \cite{mees2022calvin}. In the Language-table dataset, a robot gripper rearranged tabletop objects following language commands. The data were collected through teleoperation in a simulation, totaling 4.93 million frames. In the Calvin dataset, a 7-DOF robot manipulator performed manipulation tasks following relatively abstract instructions linked with a series of language commands. We utilized only the data containing language instructions, which amounted to 1.44 million frames. We chose these two datasets to gain insights into the model's performance across two dimensions: language-instruction abstraction and task-step length.

\begin{figure}[t]
    \centering
    \includegraphics[width=\linewidth]{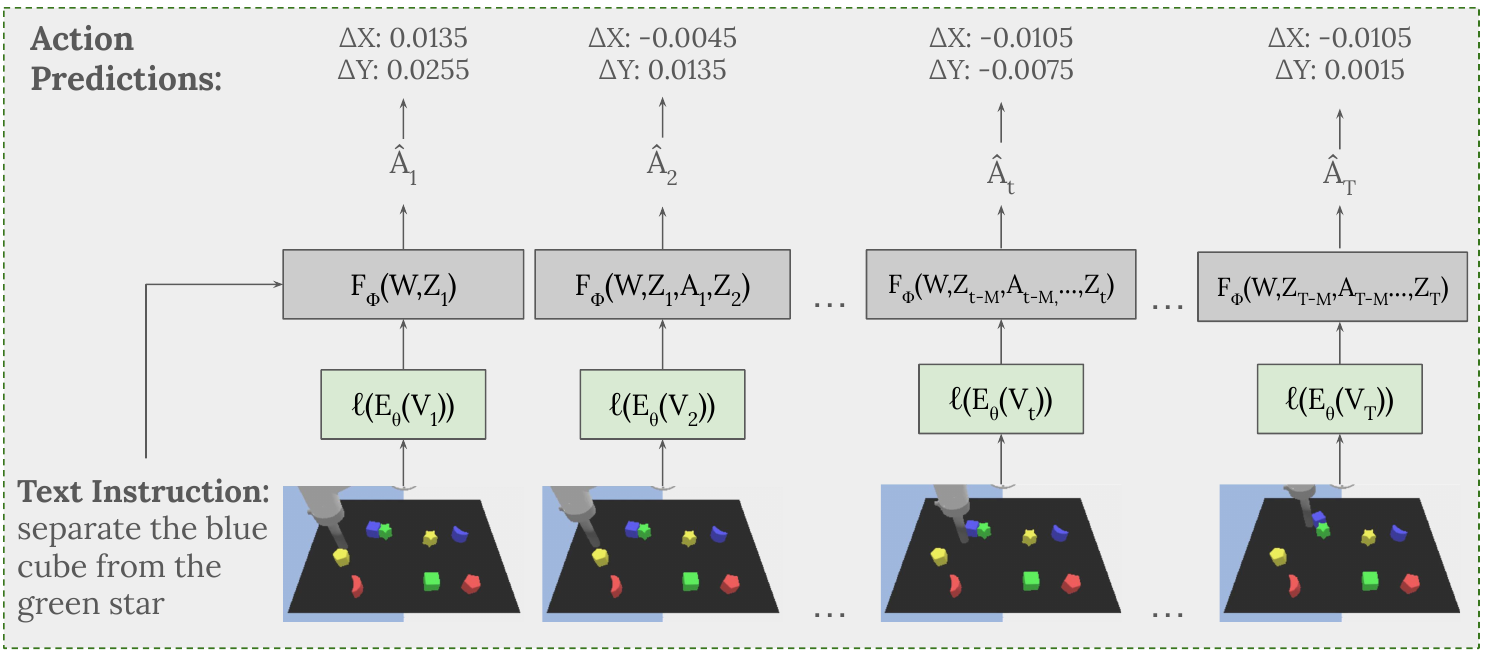}
    \vspace{-7mm}
    \caption{Our robotics and gaming pre-training pipeline.  For simplicity, we use the same notation as in Sections~\ref{sec:architecture} and \ref{sec:pretraining-strategy}; we represent our text instruction as $W$, input frames as $V_t$, our visual encoder and linear projection layer as $E_\theta$ and $\ell$, respectively, our action and language transformer model as $F_\phi$, and the predicted actions at time step $t$ as $\hat{A}_t$.}
    \label{fig:robotics_task}
\end{figure}

\subsection{Gaming Tasks}

Our primary gaming dataset consists of the Minecraft demonstrations collected by contractors in \cite{baker2022video}. In the original dataset, contractors were simply instructed to play Minecraft with no specific goal, and the dataset provided video gameplay synchronized with player actions and inventory metadata. However, since our architecture can leverage text instructions, we use GPT-4V to label videos with more specific instructions. Our prompt to GPT-4V also includes changes in the player's inventory over the video, which we found helped to reduce misclassifications of objects and actions in the video. In total, the Minecraft portion of our pre-training dataset consists of 4.7 million frames.

In addition to Minecraft, we also used a dataset of gameplay from Bleeding Edge, a team-base multiplayer game, which consists of video and synchronized player actions. Similarly, there are no specific instructions provided with the video, so we use GPT-4V to label the videos in our dataset. The Bleeding Edge portion of our pre-training dataset consists of 2.3 million frames across 7 different settings in the game.

\subsection{Healthcare Tasks}

In the healthcare domain we explored, our main dataset consisted of real-world recorded scenes from hospital ICU (intensive care unit) rooms using wall-mounted RGB cameras. Experienced ICU nurses generated captions of extracted 5-10 second video clips depicting common nursing activities in the ICU. We also included routine nursing documentation of important observations based on longer 5-30 minute windows, which included common clinical measures that assist with assessment and treatment of the patient’s condition. For the analysis described in this paper, we focused on the RASS (Richmond Agitation-Sedation Scale) score used to assess the patient’s state of agitation and sedation \cite{sessler2002richmond} and the bed position to confirm that the head of the bed is at the proper angle to decrease the chance of acquiring a ventilator-associated pneumonia \cite{keeley2007reducing}. Both assessments are recorded frequently in the medical record and automated documentation has the potential to optimize caretaker time.

\begin{figure}
    \centering
    \includegraphics[width=\linewidth]{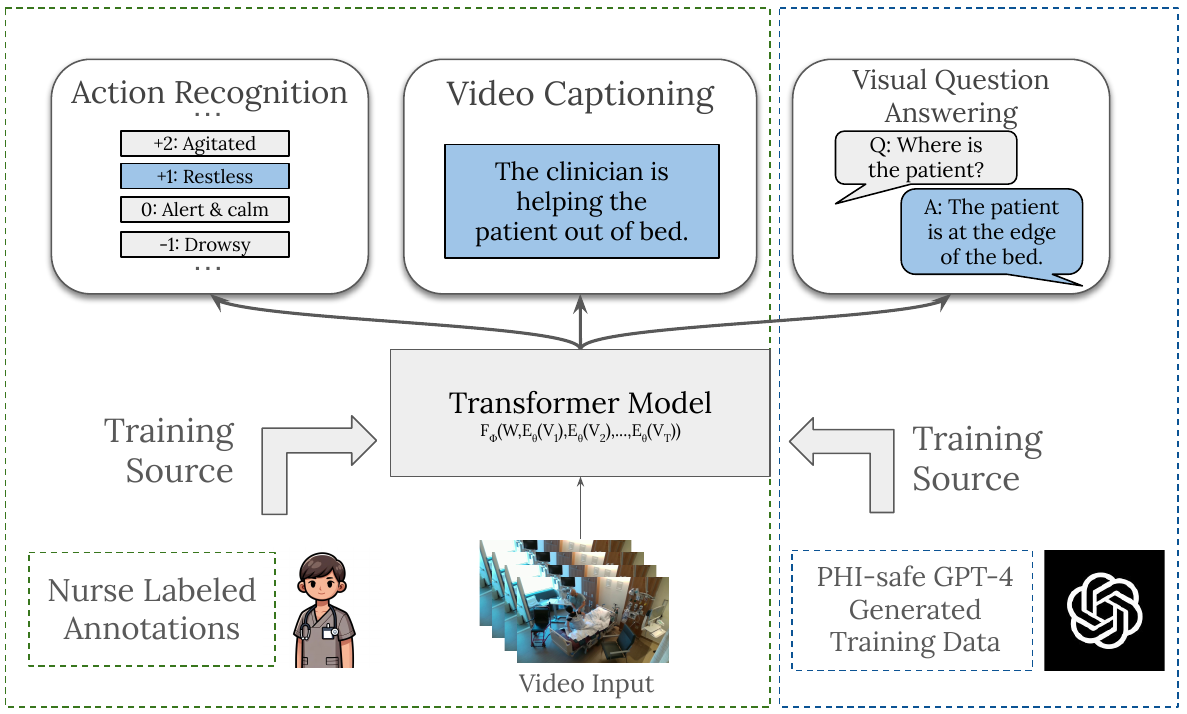}
    \vspace{-8mm}
    \caption{A High-level Overview of our Healthcare Tasks. We leveraged nurse-labeled annotations to train our multimodal agent on healthcare data.  To adapt our model for visual question answering, we generated additional training data with GPT-4 using the PHI-safe process shown in Appendix~\ref{sec:gpt-4-prompting}.}
    \label{fig:hospital_tasks}
    \vspace{-0.5cm}
\end{figure}

In order to fine-tune our model for human interactions in our ICU use case, we leveraged the nurse-provided video-clip captions and clinical documentation to have GPT-4 generate a synthetic video question-answer dataset that was used to expand the capabilities of our model after healthcare fine-tuning. A definite advantage of the GPT-4 generated derivative dataset is that it did not use any confidential patient data and consequently can be made publicly available to train any language-grounded clinical model. Figure~\ref{fig:hospital_tasks} provides an overview of the healthcare tasks we evaluated: (1) video captioning, (2) video question answering, and (3) RASS score prediction (which we formulate as an activity recognition problem). For more information about our GPT-4 based question-answer generation procedure, see Appendix \ref{sec:gpt-4-prompting}.



\vspace{0.5cm}
\section{Experiments}
\label{sec:experiments}

From a technical perspective, we are developing a generic artificial intelligence agent foundation model that can understand a wide array of input modalities and can produce coherent outputs and actions within a wide range of diverse interactive environments.  
In addition to evaluating our framework in these more specific domains, we evaluated the capabilities of our pre-training model on robotics manipulation, game playing, and interactive healthcare tasks. The details of the experimental setting and our main results are described in the following sub-sections.

\subsection{Pre-training Experiments}
\label{sec:pretraining}

To pre-train our model, we used the full training sets of Language Table, CALVIN, Minecraft, and Bleeding Edge, and trained for 100 epochs. We used a linear warmup cosine learning rate scheduler, with an initial learning rate of 0.0001. We initialized the vision component of our model with the CLIP base model with patch size 16, and initialized the language and action components with OPT-125M. We used 12 nodes of 16 V100 GPUs for 175 hours for all of our pre-training.

We added new action tokens corresponding to the actions used in our training set. All tasks include a token to indicate starting actions and a token to indicate ending actions. For Minecraft, there are additionally 23 button actions, and we discretized mouse actions to 100 bins along the x axis and 100 bins along the y axis. For Bleeding Edge, there are 11 button actions, and 2 joysticks. Each joystick has 256 possible values for rotation and 4 values for magnitude, resulting in a total of 520 joystick action tokens.

For robotics, we added new action tokens corresponding to valid actions in the environment, along with agent state tokens for proprioception. For all robotics data, we included a special action token to indicate the end of a trajectory. In Language Table, we included 21 binned actions for each of the $x$ and $y$ directions, representing the end effector translation target. We also included 21 binned state tokens representing the current end effector translation for each of the $x$ and $y$ directions, and an equal number of state tokens representing the previous robot action. In CALVIN, we included two actions for the gripper, indicating opening and closing, along with 21 actions for each of the six degrees of freedom of the end effector in the relative Cartesian displacement action space. We also included 21 binned states for each of the 14 attributes of the proprioceptive state, excluding the gripper action which has two states.

\begin{figure}
    \centering
    \includegraphics[width  =0.95\linewidth]{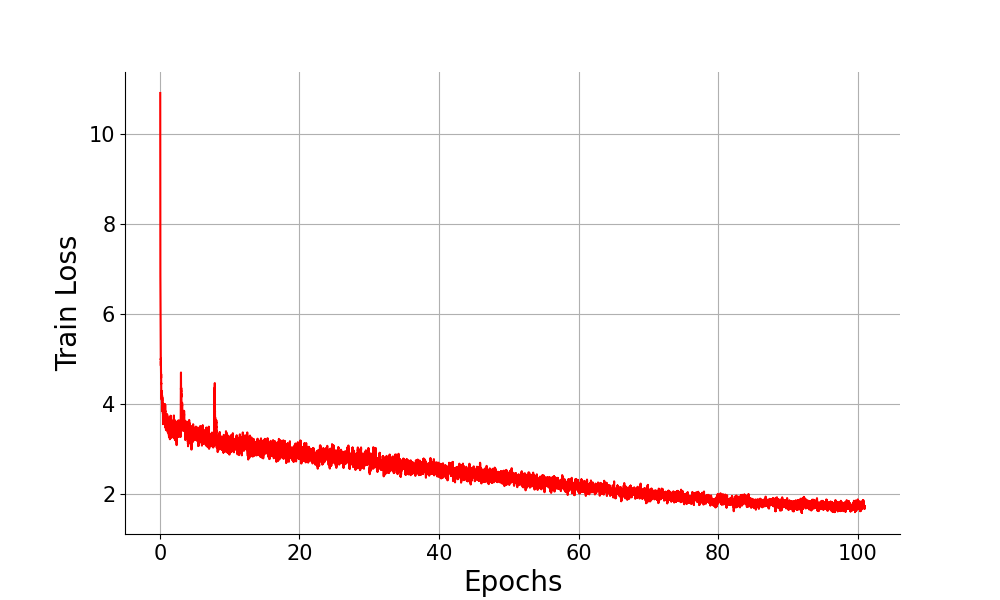}
    \caption{Plot of total pre-training loss over 100 epochs.}
    \label{fig:loss_plot}
\end{figure}

Our gaming dataset has 525,309 trajectories for Minecraft and 256,867 for Bleeding Edge, each consisting of 9 frames. Our robotics dataset consists of 1,233,659 trajectories for Language-Table and 360,566 for CALVIN, each consisting of 4 frames. Therefore, our total dataset consists of 13,416,484 frames. When sampling trajectories to train our model, we additionally added color jitter to each of the images, randomly scaling the brightness and saturation between 70\% and 140\%, and randomly shifting the hue by at most 0.05. We plot our pre-training loss in Figure \ref{fig:loss_plot}.

\begin{table*}[ht]
\caption{Results for robotics fine-tuning across tasks on CALVIN and Language-Table, along with their corresponding evaluation metrics.
 }
\vspace{-2mm}
\label{tab:calvin_baselines}
\vskip 0.15in
\begin{center}
\begin{scriptsize}
\begin{sc}
\resizebox{0.85 \linewidth}{!}{%
\begin{tabular}{lcccccc|c}
\toprule
& & & CALVIN & & & & Language Table \\
\midrule
\small{Model}  &  \small{1 step} & \small{2 step} & \small{3 step} & \small{4 step} & \small{5 step} & \small{Avg Lens} & \small{Success Rate}\\
\midrule
\small{MCIL} & \small{28.2} & \small{2.5} & \small{0.3} & \small{0.0} & \small{0.0} & \small{0.31} & \small{---}\\
\small{Ours} \tiny{(From Scratch)} & \small{20.6} & \small{0.8} & \small{0.0} & \small{0.0} & \small{0.0} &\small{0.214} & \small{40.0} \\
\small\textbf{Ours} & \small\textbf{64.8} & \small\textbf{29.0} & \small\textbf{12.3} & \small\textbf{4.7} & \small\textbf{1.9} & \small\textbf{1.127} & \small\textbf{42.0}\\
\bottomrule
\end{tabular}
}
\end{sc}
\end{scriptsize}
\end{center}
\vspace{-5mm}
\end{table*}

\subsection{Robotics Experiments}

The pre-trained model was fine-tuned for the Language-Table and CALVIN datasets and evaluated separately. 
For fine-tuning, we used the same pipeline as in pre-training, maintaining the original MAE and language-modeling loss functions, and the original vocabulary size. During fine-tuning, 50\% of the image patches were masked, while no masking was involved in the evaluation.

\subsubsection{Language-Table}

In the Language-table dataset, we used data from a setup involving a total of 8 blocks, out of which 6 blocks were non-manipulated and unrelated to the tasks. This setup resulted in 181,020 trajectories. 
We split each trajectory into a series of 4 frames to fit our model architecture, resulting in 1,233,659 samples for fine-tuning. To investigate performance against different task characteristics, the model was evaluated on 5 different subtasks: 1) moving a block to another block; 2) moving a block relative to another block; 3) moving a block to an absolute position; 4) moving a block to a relative position; 5) separating two blocks. For each task,  50 trajectories were randomly sampled and evaluated three times, and the average success rate was computed. 
While the pre-trained model performed better than training from scratch (Table~\ref{tab:calvin_baselines}), our model was outperformed by other models such as \cite{brohan2023rt}, which could be attributed to the fact that we used less data for pre-training, only using the human-teleoperated data in the Language-Table, CALVIN, and gaming datasets.

\subsubsection{CALVIN}

In the CALVIN dataset, each long-step trajectory was split into a series of 4 frames, resulting in 360,566 samples across 34 tasks for fine-tuning. 
To better capture the entire scene, the third-person view RGB camera was chosen as the source of image input from the available camera resources. 
For fine-tuning, we incorporated all available appearance settings, including the one used for testing, to enlarge the dataset, following the standard $ABCD \rightarrow D$ task definition. To evaluate the model performance with multiple steps, we computed the averaged success rate at each step, following the methodology described in the original CALVIN paper~\cite{mees2022calvin}. 
Compared to Multi-context Imitation Learning (MCIL) \cite{lynch2021language}, our model shows better performance while only using 1\% of the data (Table~\ref{tab:calvin_baselines}).

\vspace{1mm}
\begin{table}[t]
\caption{Performance metrics for gaming data. We report BLEU-4 scores for action prediction in Minecraft (abbreviated as MC), and Bleeding Edge (abbreviated as BE). We choose the last epoch for the pre-trained model and the epochs with the best validation score for the other models.}
\vspace{-5mm}
\label{tab:gaming_baselines}\vskip 0.15in
\begin{center}
\begin{scriptsize}
\begin{sc}
\begin{tabular}{lccr}
\toprule
Model & MC (BLEU-4)$\uparrow$ & BE (BLEU-4)$\uparrow$ \\
\midrule
Ours (from scratch) & 0.174 & 0.238 \\
Ours (pre-train only) & 0.170 & 0.249 \\
Ours (Pre-train and fine-tuned) & \textbf{0.272} & \textbf{0.411} \\
\bottomrule
\end{tabular}
\end{sc}
\end{scriptsize}
\end{center}
\vspace{-5mm}
\end{table}

\begin{table*}[ht!]
\centering
\begin{tabular}
{@{}m{1.5cm}|@{}m{3.3cm}|c|@{}m{3.3cm}|@{}m{3.3cm}}
\centering
Task & Text instruction & Start frame & Predicted Action &  Ground Truth Action \\
\toprule
Minecraft & the player is using an iron\_sword to attack and kill pigs in a forest... & 
\includegraphics[width=0.22\textwidth,valign=m]{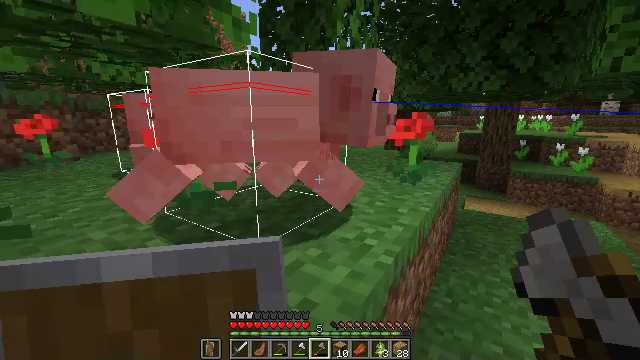} &
[STARTACTION] [attack] [ENDOFACTION] & 
[STARTACTION] [attack] [ENDOFACTION] \\
\addlinespace[2.5pt]
\midrule
Bleeding Edge & the player is controlling a red robot ... fighting other characters & 
\includegraphics[width=0.22\textwidth,valign=m]{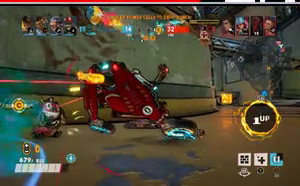} &
[STARTACTION] [lockon][meleeattack] [lrot162] [lmag4] [ENDOFACTION] & 
[STARTACTION] [lockon][meleeattack] [lrot160] [lmag4] [ENDOFACTION] \\
\addlinespace[2.5pt]
\bottomrule
\end{tabular}
\caption{Examples of actions predicted by our fine-tuned models for Minecraft (above) and Bleeding Edge (below). More examples are presented in Appendix~\ref{sec:appendix_examples}.}
\label{tab:gaming_samples}

\end{table*}

\subsection{Gaming Experiments}

For both gaming settings of Minecraft and Bleeding Edge, we evaluated our model's ability to predict actions given video frames and high-level instructions, along with its MAE reconstruction quality. Specifically, we used a held-out test dataset of 100 videos each, formatted in the same manner as our training data.

We report the BLEU-4 scores of actions in Table~\ref{tab:gaming_baselines}. We compare our pre-trained baseline to fine-tuning on task-specific data initialized from our pre-trained model and a version initialized from CLIP and OPT. We find that both fine-tuned models over-fit to the training data within 5 epochs, so we report the BLEU-4 test scores from the checkpoints with the highest validation score. We find that fine-tuning our pre-trained model is significantly more effective than training from scratch for both gaming domains, highlighting the importance of our diverse pre-training mixture. We also show a visualization of predicted actions from our fine-tuned model compared to the validation ground-truth in Table~\ref{tab:gaming_samples} and Appendix~\ref{sec:appendix_examples}. 

\subsection{Healthcare Experiments}

For our experiments on our healthcare dataset, we evaluated our model's ability on three separate downstream tasks: video captioning, visual question answering, and activity recognition in the form of RASS score prediction.  We used the final checkpoint from our pre-training run as described in Section~\ref{sec:pretraining}.

\vspace{-3mm}
\paragraph{Healthcare Setting} 
\label{sec:healthcare_baselines}

For visual question-answering, we use the question as the text prompt $W$, and use the fixed text prompt ``A video of" for video captioning.  We train our model to the corresponding text tokens of the caption or answer and report the average perplexity across both settings.  We frame RASS score prediction as a 10-way activity classification problem, and train a separate classification head for our model.  We use the video-level setting for our visual encoder with 9 frames as input, as described in Appendix \ref{sec:appendix_architecture}. To evaluate the effectiveness of our pre-training framework, we compared the performance of our model against three baselines that leverage CLIP and OPT for initialization.    First, we compared against a \textit{frozen} baseline that uses the same pre-trained models, kept frozen, while fine-tuning a single linear layer for cross modal information passing, similar to \cite{liu2023llava}.  Second, we compared against a \textit{joint} baseline that uses the same pre-trained models but fine-tunes them jointly along with the linear layer.  For both of these baselines, we encode frames with CLIP individually and concatenate the frame-level embeddings. Third, we compared against a baseline of our same architecture, that makes use of our video-level encoder and is initialized from CLIP and OPT, but does not use any large-scale agent pre-training.  We show our performance against the proposed baselines in Table~\ref{tab:healthcare_baselines}.  For all results, we train for 20 epochs on 4 16GB V100 GPUs with a fixed learning rate of 4e-5 and report results on a held-out evaluation set.  For fair comparison, we do not perform any additional hyperparameter search.

\begin{table}[t]
\caption{Performance on healthcare text generation and RASS score action recognition, along with the corresponding evaluation metrics. Agent pre-training on robotics and gaming data improves performance for action recognition, but does not improve text generation abilities.}
\label{tab:healthcare_baselines}
\vskip 0.15in
\begin{center}
\begin{scriptsize}
\begin{sc}
\begin{tabular}{lcc}
\toprule
Model & Perplexity $\downarrow$ & RASS Acc $\uparrow$ \\
\midrule
CLIP + OPT (frozen) & \textbf{93.3} & 55.4 \\ 
CLIP + OPT (unfrozen) & 102.7 & 92.6 \\
Ours (from scratch) & 100.0 & 70.3 \\
Ours (Agent pre-trained) & 106.3 & \textbf{95.7} \\
\bottomrule
\end{tabular}
\end{sc}
\end{scriptsize}
\end{center}
\vspace{-7mm}
\end{table}

\section{Ablations and Analysis}
\label{sec:discussion}

\paragraph{Pretraining Loss Curves:} We plot our combined pre-training loss across 100 epochs in Figure~\ref{fig:loss_plot}, and show individual components of the loss function in Appendix \ref{sec:appendix-loss-curves}.
\vspace{-4mm}

\paragraph{Comparisons with GPT-4V:} In Figure~\ref{fig:gpt4v-comparison}, we show how our model has the ability to output low-level action predictions, while GPT-4V is unable to consistently output low-level controls.  While our model is able to output precise movements and actions, GPT-4V only outputs high-level instruction.
\vspace{-4mm}

\paragraph{Effects of Agent Pre-Training:}  In Table~\ref{tab:gaming_baselines} and Table~\ref{tab:healthcare_baselines}, we demonstrate the effectiveness of our agent pre-training strategy compared to training from scratch and training against an equivalent visual-language baseline. In particular, we show that a commonly used approach for fine-tuning visual-language models by using frozen visual encoders, similar to LLaVA \cite{liu2023llava} or Mini-GPT-4 \cite{zhu2023minigpt4}, performs worse than joint fine-tuning for action recognition on our healthcare dataset.  Furthermore, our agent pre-training boosts performance for action prediction across all gaming and robotics datasets. 


\section{Conclusion}
\label{sec:conclusion}

We introduced an Interactive Agent Foundation Model designed to take text, action, and visual inputs. We found that by pre-training on a mixture of robotics and gaming data, our model is effective in modeling actions across a variety of domains, even showing positive transfer when fine-tuning in unseen domains such as healthcare. The generality of our framework allows it to be broadly applicable across decision-making settings, unlocking new possibilities for generalist agents in multimodal systems.


\section{Impact Statement}

This paper presents the initial steps on making interactive agents possible through an Interactive Agent Foundation Model. We do not foresee negative societal consequences from presenting and open-sourcing our current work. In particular, the main output of our model is domain-specific actions, such as button inputs for gaming data, making the downstream applications of our model different from those of standard LLMs and VLMs. 

In the domain of robotics, we wish to emphasize that our model should not be deployed on real robots without more training and additional safety filters.

In the domain of gaming, downstream applications of our foundation model may have some societal consequences. Smarter, more realistic AI characters could lead to more immersive worlds, which can increase players' enjoyment in games, but may also lead to social withdrawal if not used appropriately. Specifically, more realistic AI characters could potentially lead to video game addiction and players anthropomorphising artificial players. We encourage game developers who build AI agents using our models to mitigate these potential harms by encouraging social interactions between human players and applying appropriate content filters to AI agents.

In the domain of healthcare, we emphasize that our models are not official medical devices and have not gone through rigorous testing in live settings. We strongly discourage using our models for self-prescription. Even as our models improve in future iterations, we strongly encourage keeping a medical practitioner in the loop to ensure that unsafe actions are avoided. As our models continue to develop, we believe that they will be useful to caretakers, especially by automatically forming drafts of documentation and notifying caretakers when patients may need urgent attention.

Finally, we note that the capabilities of agent AI models may significantly change at scale.  As we scale our model in terms of architecture, compute, and training data, we will actively monitor its capabilities before releasing new versions publicly.

\section*{Acknowledgements}

We are especially grateful to Desney Tan, Peter Lee, Doug Burger, Ryen White, Ece Kamar, 
John Langford, Jonathan Carlson and Microsoft's Office of the CTO (OCTO) for their advice, enormous support, and encouragement. We appreciate the Microsoft gaming team, Microsoft X-box team, Microsoft 343 team, Kareem Choudhry, Haiyan Zhang, Spencer Perreault, Dave Bignell, Katja Hofmann, Sam Devlin, Shanzheng Tan, and Raluca Georgescu for the gaming data collection and sharing. We thank Bill Dolan, Nebojsa Jojic, Sudha Rao, Adrian Brown, Andrzej Banburski-Fahey, and Jianwei Yang for their early insightful discussions and help with the gaming aspects of our project. We appreciate Kiran Muthabatulla and the MSR Central Engineering (CE) team for their discussion and feedback for the project. The authors gratefully acknowledge the Microsoft HoloLens team, Microsoft Mesh team, and Antonio Criminisi for their generous provision of equipment and project discussions. Finally, we would like to express our genuine appreciation for Jim Jernigan, Ben Huntley, Oleg Losinets, the Microsoft AOAI team, and the GCR team for their Azure-OpenAI endpoint support and their pointers to the literature. 

We would also like to thank our colleagues from Stanford's Partnership in AI-assisted Care, who helped inform the medical applications explored in this work.  In particular, we would like to thank Amit Kaushal and Roger Bohn for their clinical expertise and guidance. Additionally, we greatly appreciate Zelun Luo, David Dai, and Dev Dash for their participation as actors for our hospital dataset.

This research was supported by Microsoft Research Project Green 2024, Microsoft Research Project Fair 2023, Stanford University, University of California at Los Angeles, MSR Accelerator team, and the Microsoft OCTO team.


\bibliography{main}
\bibliographystyle{icml2024}

\clearpage
\onecolumn
\appendix
\newpage 
\vspace*{-8mm} 
\begin{center}
    \LARGE{\textbf{Appendix}}
\end{center}
\section{Architecture Details}
\label{sec:appendix_architecture}

To effectively handle images and video inputs jointly, we use a divided space-time attention similar to \cite{bain2021frozen}.  We initialize our visual encoder from CLIP ViT-B16 \cite{radford2021learning}, and learn temporal attention layers after each spatial attention layer.  We further mask 75\% of the image patches (using tubelet masking for videos) during training, and use a MAE-decoder similar to \cite{he2021masked}. Gaming and robotics use a frame-level visual encoder so that the agent is able to observe a continuous stream of tokens and act after every frame.  For healthcare, we leverage the video understanding capabilities of our visual encoder since the tasks are video-level.

\section{GPT-4 Prompting}
\label{sec:gpt-4-prompting}

\begin{figure}[ht]
    \centering
    \includegraphics[width=0.5\linewidth]{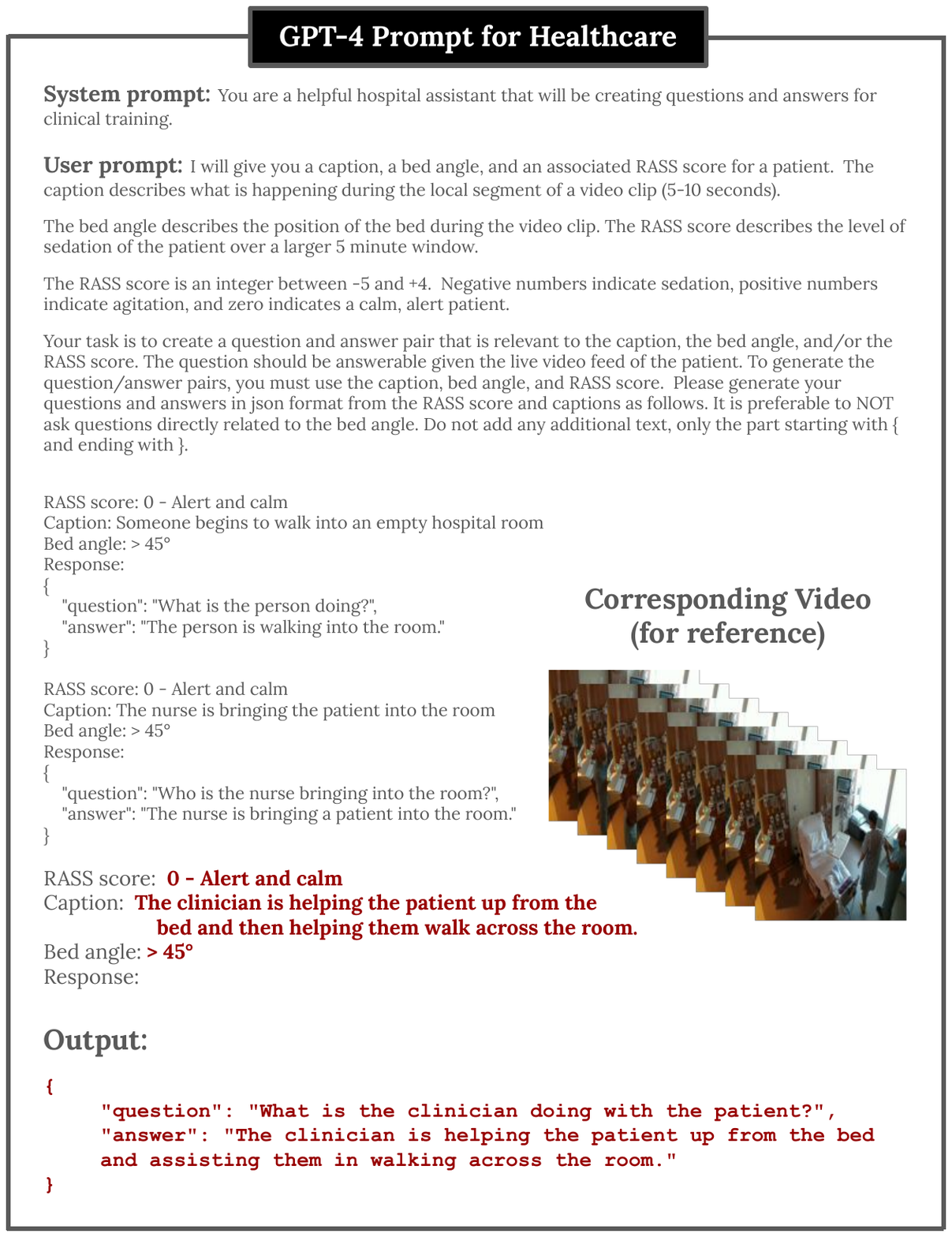}
    \caption{\textbf{Our PHI-safe GPT-4 Prompt for Generating Healthcare QA Examples.}  By ensuring the usage of non-identifying video captions and documentation data, we prevent any identifiable patient data leakage to GPT-4 while simultaneously generating additional visual-language training data.  For the particular example shown, we use a RASS score of ``0 - Alert and calm", a caption of ``The clinician is helping the patient up from the bed and then helping them walk across the room.", and a bed angle of `` $>45$°".}
        \label{fig:hospital_qa}
\end{figure}

\begin{figure}[ht]
    \centering
    \includegraphics[width=0.5\linewidth]{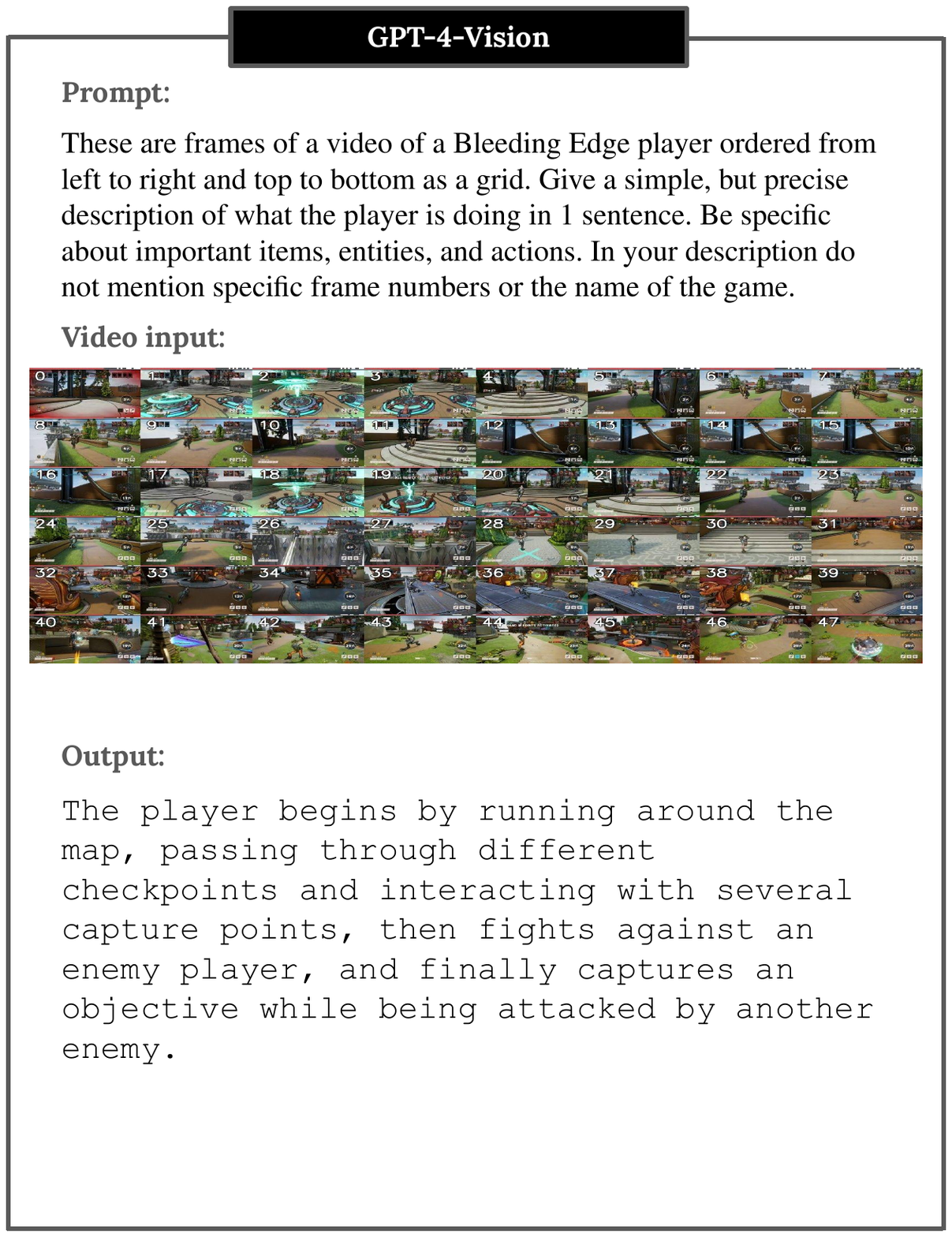}
    \caption{Our GPT-4V prompt for games like Bleeding Edge that have 3rd person viewpoints and visually complex scenes. In order to input a large number of frames (48) to GPT-4V, we input the frames as a grid with frame numbers overlaid on each frame (as shown above).}
    \label{fig:bleeding_gpt4}
\end{figure}

\begin{figure}[ht]
    \centering
    \raisebox{-0.5\height}{\includegraphics[width=0.45\linewidth]{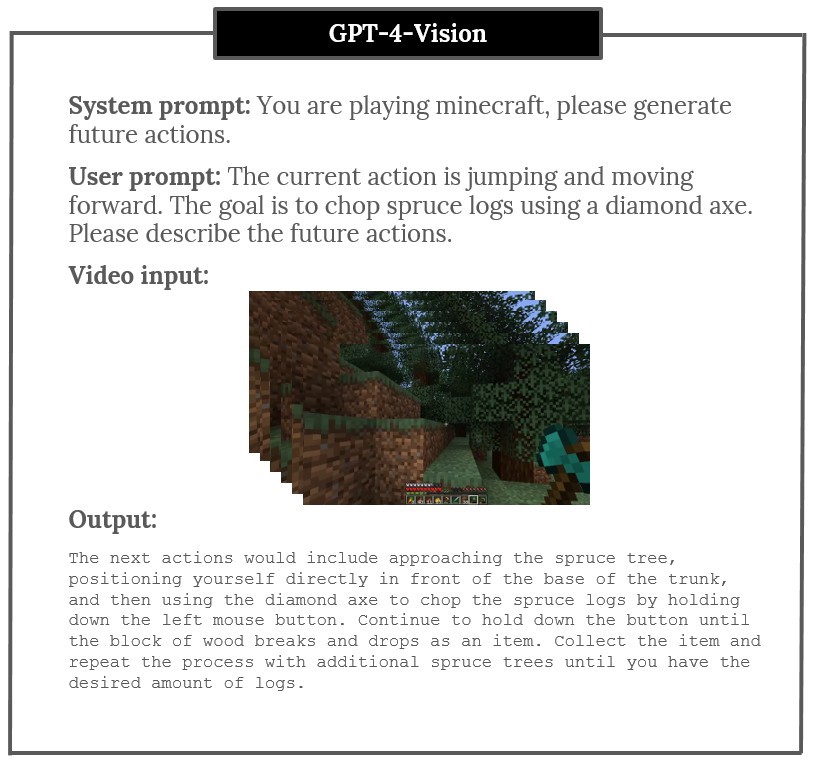}}
    \raisebox{-0.5\height}{\includegraphics[width=0.45\linewidth]{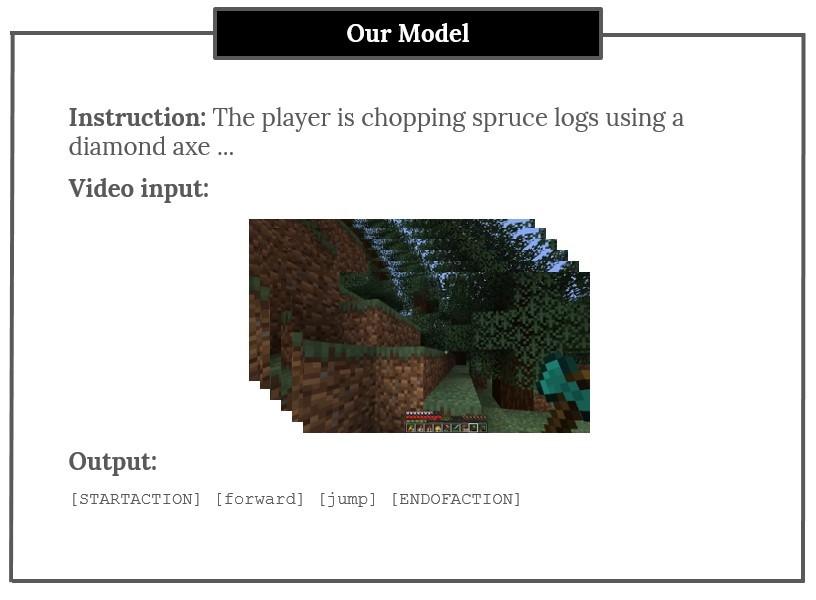}}
    \caption{When using GPT-4V to choose actions given a history of frames, we find that it gives reasonable high-level actions but does not choose precise low-level actions, highlighting the importance of our pre-trained model.}
    \label{fig:gpt4v-comparison}
\end{figure}

We show our GPT-4 Prompt for Healthcare Visual Question Answering generation in Figure \ref{fig:hospital_qa}, and our GPT-4V Prompt for gaming instruction generation in Figure \ref{fig:bleeding_gpt4}. 

\section{Pre-training Loss Curves}
\label{sec:appendix-loss-curves}

\begin{figure}[ht]
\centering
\begin{tabular}{cc}
  \includegraphics[width=65mm]{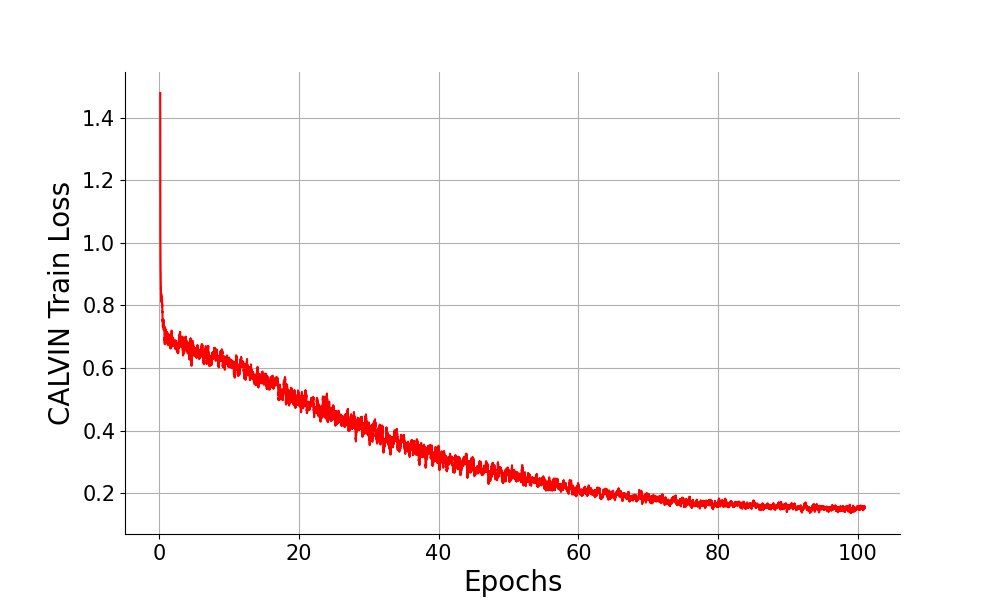} &   \includegraphics[width=65mm]{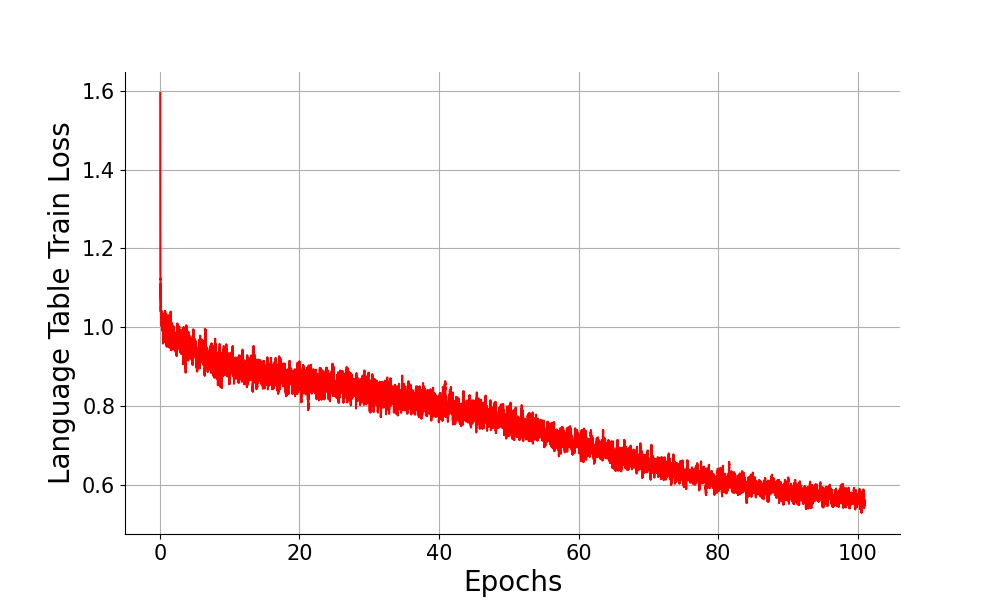} \\
 \includegraphics[width=65mm]{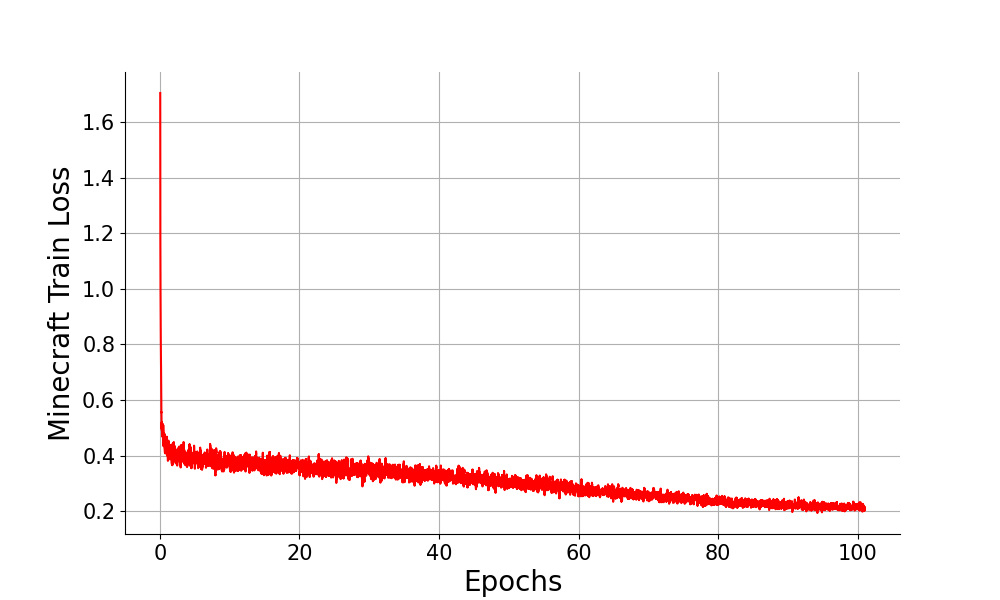} &   \includegraphics[width=65mm]{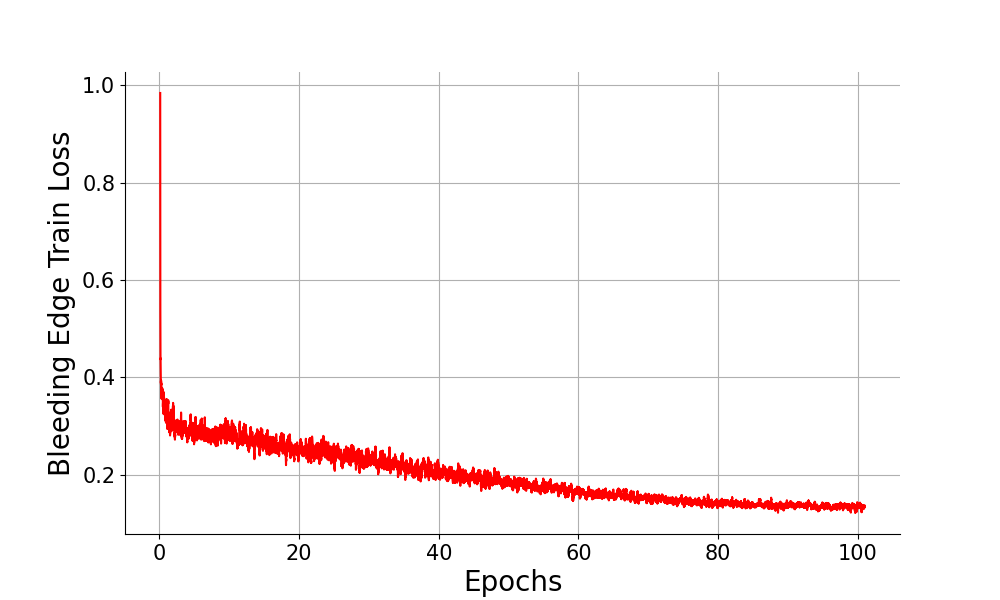} \\
\multicolumn{2}{c}{\includegraphics[width=65mm]{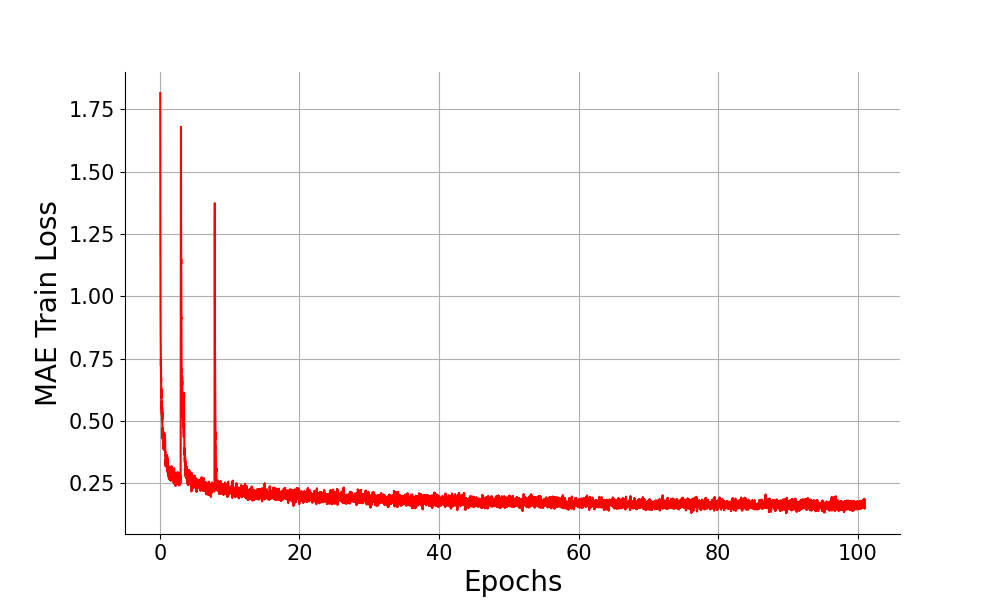} }\\
\end{tabular}
\caption{Plot of all components of the training loss over the 100 epochs of pre-training.}
\label{fig:all_losses}
\end{figure}

We show all components of the loss function in Figure~\ref{fig:all_losses} and plot our combined pre-training loss across 100 epochs in  Figure~\ref{fig:loss_plot}.

\section{Gaming Task Pipeline}

\begin{figure}[ht]
    \centering
    \includegraphics[width=\linewidth]{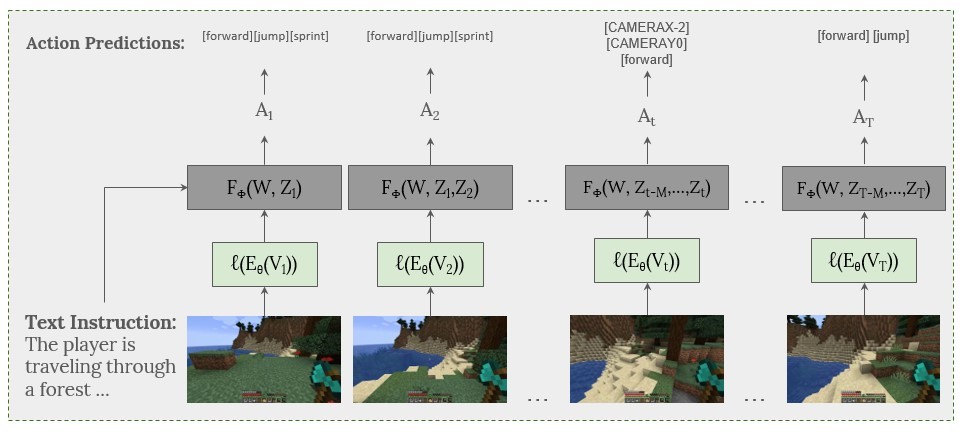}
    \vspace{-7mm}
    \caption{Our gaming pre-training pipeline.  For simplicity, we use the same notation as in Sections \ref{sec:architecture} and \ref{sec:pretraining-strategy}; we represent our text instruction as $W$, input frames as $V_t$, our visual encoder and linear projection layer as $E_\theta$ and $\ell$, respectively, our action and language transformer model as $F_\phi$ and the predicted actions at time step $t$  as $\hat{A}_t$.   }
    \label{fig:gaming_task}
\end{figure}

We provide an example of our pipeline for a gaming task in Figure~\ref{fig:gaming_task}. Note the similarities to the robotics task in Figure~\ref{fig:robotics_task} since both tasks require predicting an action given a text instruction and sequence of prior actions.

\section{Example Outputs}
\label{sec:appendix_examples}

We show examples of our model predicting actions on unseen, robotics simulation data in Table \ref{tab:lt_samples} and \ref{tab:calvin_samples}.  We show example outputs for healthcare in Table \ref{tab:healthcare_samples}, and show example outputs for gaming in Table \ref{tab:minecraft_samples} and \ref{tab:be_samples}.

\begin{table*}[ht]
\centering
\begin{tabular}{@{}m{3cm}ccccc@{}}
Text instruction & Start frame & & Middle frame & &  End frame \\
\toprule

\addlinespace[4pt] Pull the red moon apart from the blue moon. & \includegraphics[width=0.22\textwidth,valign=m]{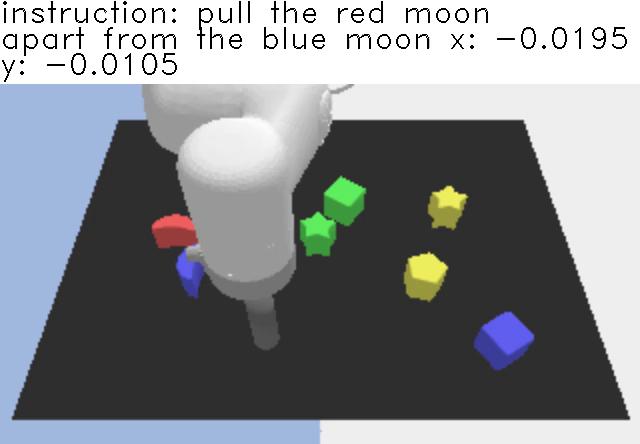} & 
$\rightarrow$ & 
\includegraphics[width=0.22\textwidth,valign=m]{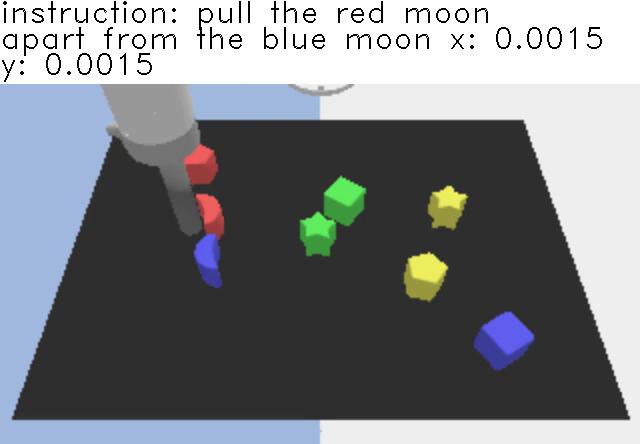} & 
$\rightarrow$ & 
\includegraphics[width=0.22\textwidth,valign=m]{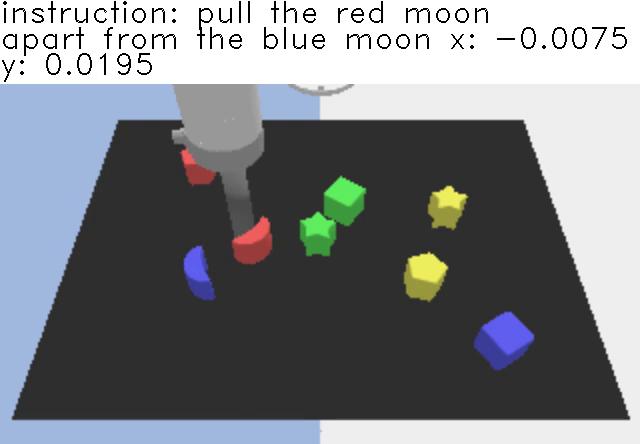} \\
\addlinespace[2.5pt]
\midrule
\addlinespace[4pt] Push the yellow start next to the red moon. & \includegraphics[width=0.22\textwidth,valign=m]{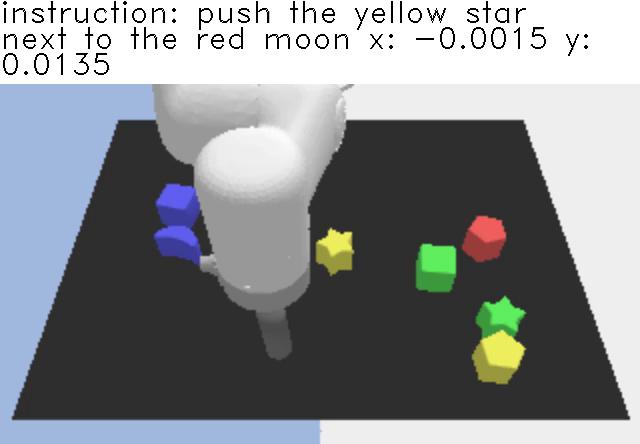} & 
$\rightarrow$ & 
\includegraphics[width=0.22\textwidth,valign=m]{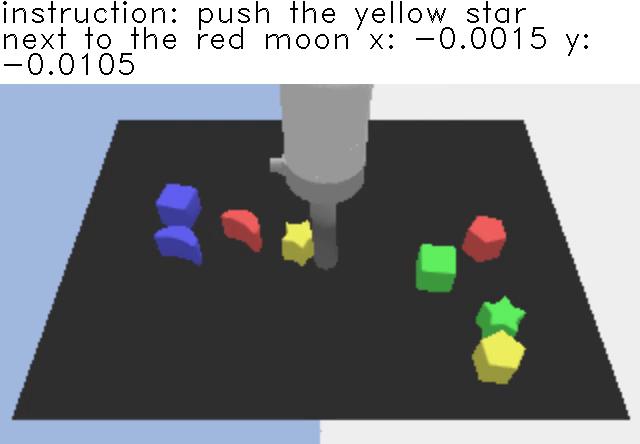} & 
$\rightarrow$ & 
\includegraphics[width=0.22\textwidth,valign=m]{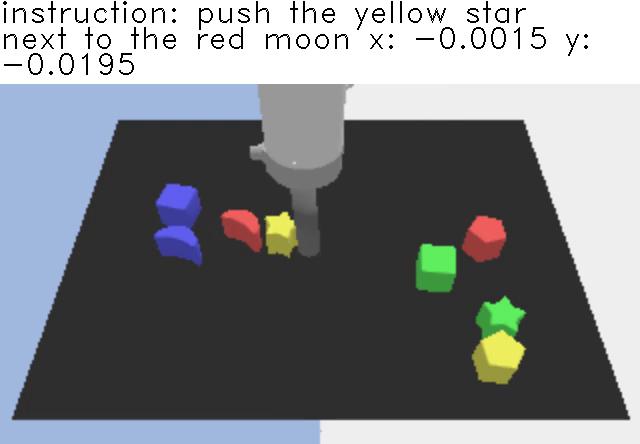} \\
\addlinespace[2.5pt]
\midrule
Move the red pentagon away from the blue cube. & \includegraphics[width=0.22\textwidth,valign=m]{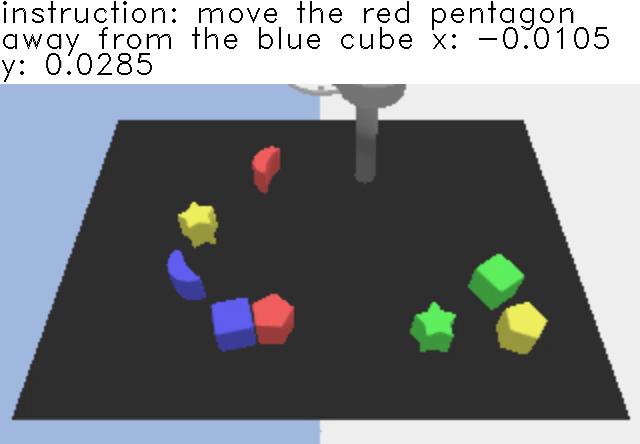} & 
$\rightarrow$ & 
\includegraphics[width=0.22\textwidth,valign=m]{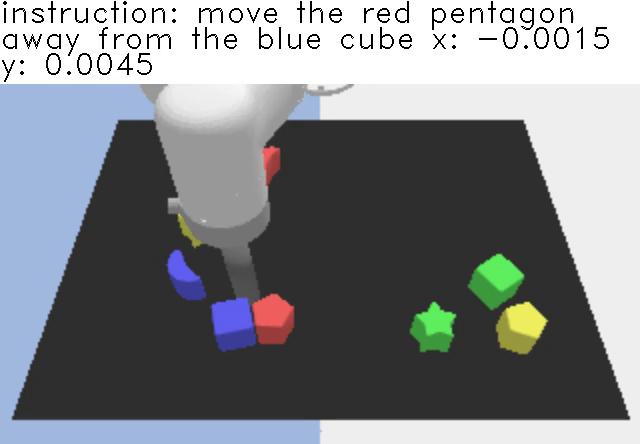} & 
$\rightarrow$ & 
\includegraphics[width=0.22\textwidth,valign=m]{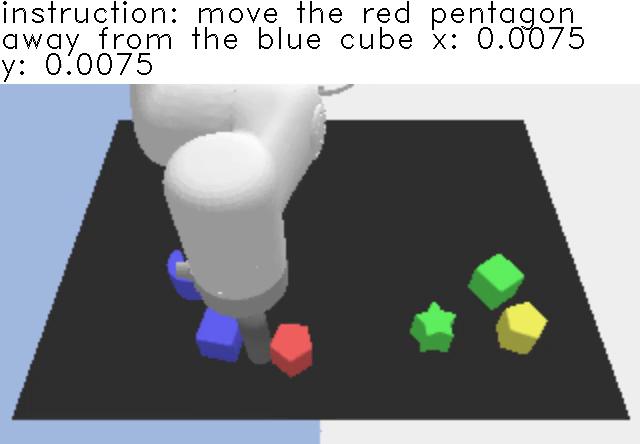} \\
\addlinespace[2.5pt]
\midrule
\addlinespace[4pt] Move the red moon to the bottom of the yellow pentagon. & \includegraphics[width=0.22\textwidth,valign=m]{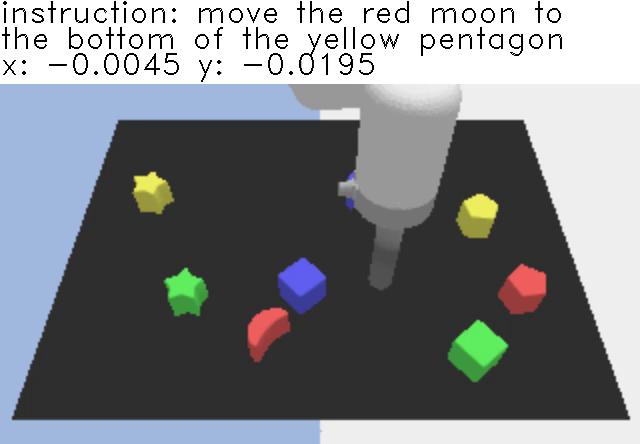} & 
$\rightarrow$ & 
\includegraphics[width=0.22\textwidth,valign=m]{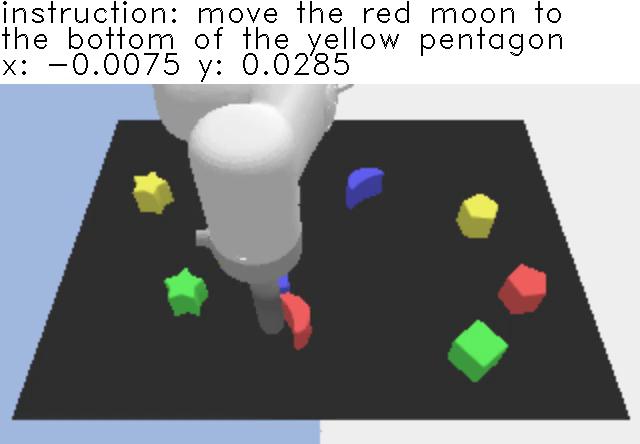} & 
$\rightarrow$ & 
\includegraphics[width=0.22\textwidth,valign=m]{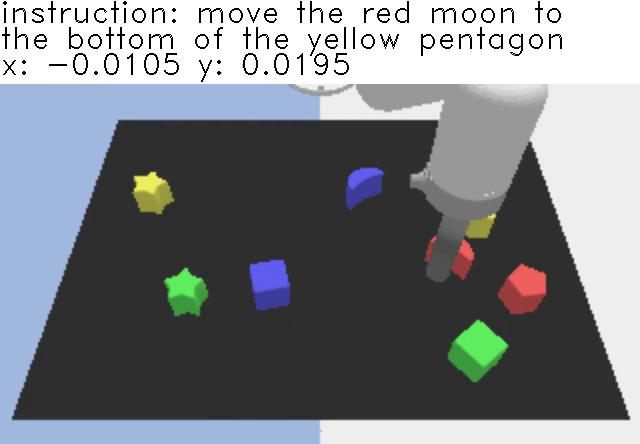} \\
\addlinespace[2.5pt]
\midrule
\addlinespace[4pt]
Pull the red moon to the bottom left. & 
\includegraphics[width=0.22\textwidth,valign=m]{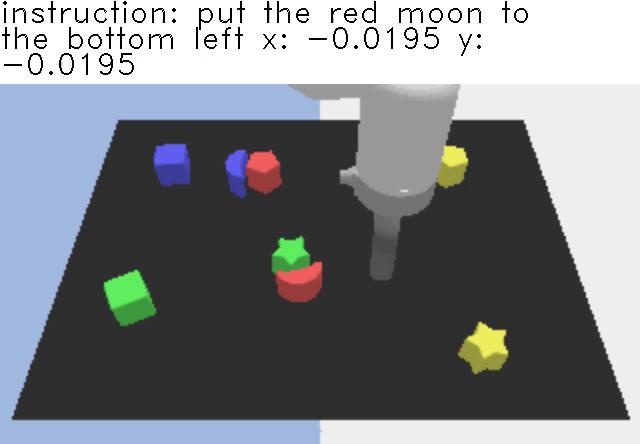} & 
$\rightarrow$ & 
\includegraphics[width=0.22\textwidth,valign=m]{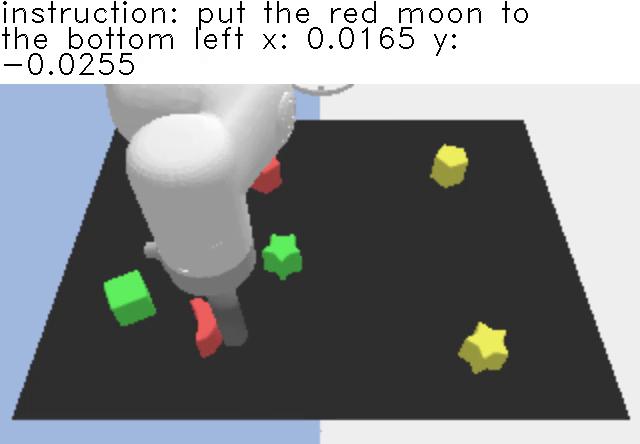} & 
$\rightarrow$ & 
\includegraphics[width=0.22\textwidth,valign=m]{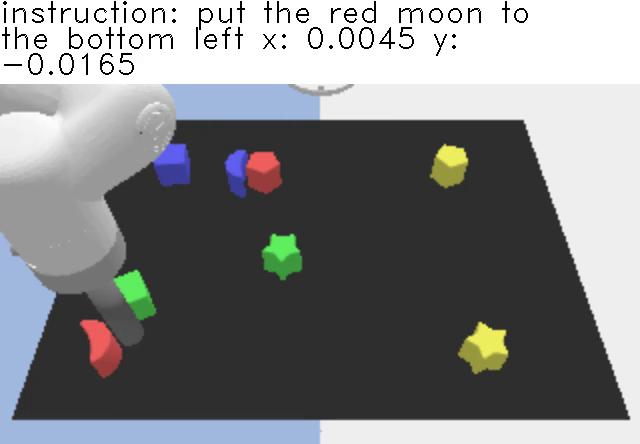} \\
\bottomrule
\end{tabular}
\caption{We show 5 unique demonstrations from Language Table, where our model successfully follows the text instruction.  In addition to the high level instruction, we also show the low-level predicted actions of our agent above each frame.}
\label{tab:lt_samples}

\end{table*}

\begin{table*}[ht]
\centering
\begin{tabular}{@{}m{3cm}ccccc@{}}
Text instruction & Start frame & & Middle frame & &  End frame \\
\toprule
Push the handle to close the drawer. & \includegraphics[width=0.22\textwidth,valign=m]{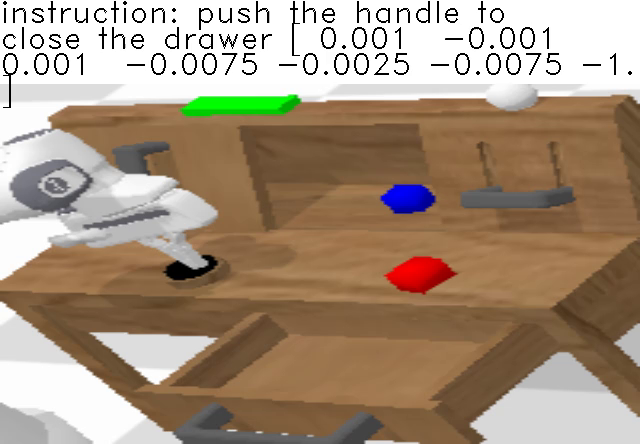} & 
$\rightarrow$ & 
\includegraphics[width=0.22\textwidth,valign=m]{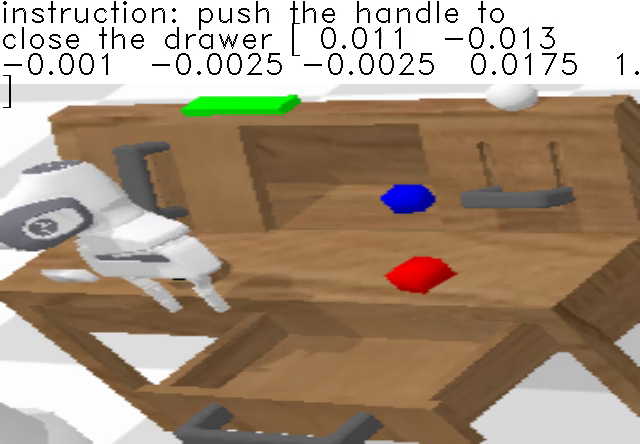} & 
$\rightarrow$ & 
\includegraphics[width=0.22\textwidth,valign=m]{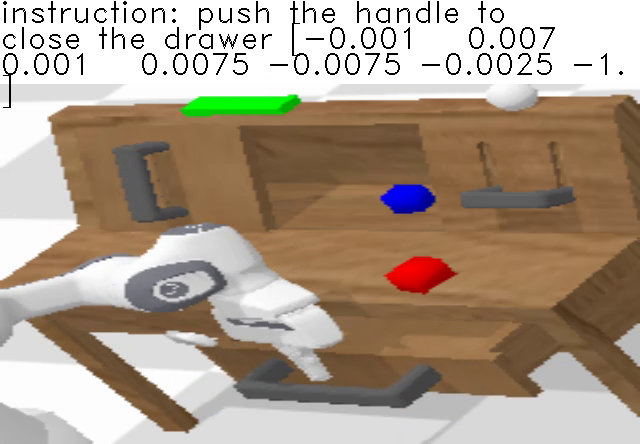} \\
\addlinespace[2.5pt]
\midrule
\addlinespace[4pt] Lift the red block from the sliding cabinet. & \includegraphics[width=0.22\textwidth,valign=m]{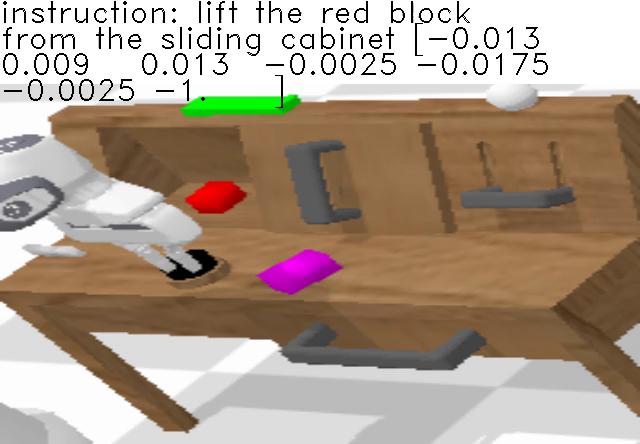} & 
$\rightarrow$ & 
\includegraphics[width=0.22\textwidth,valign=m]{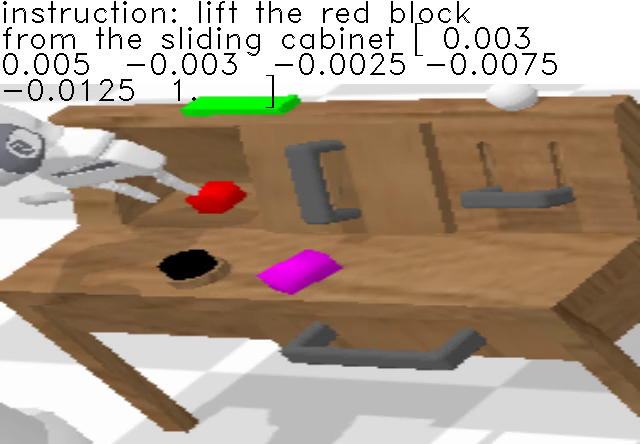} & 
$\rightarrow$ & 
\includegraphics[width=0.22\textwidth,valign=m]{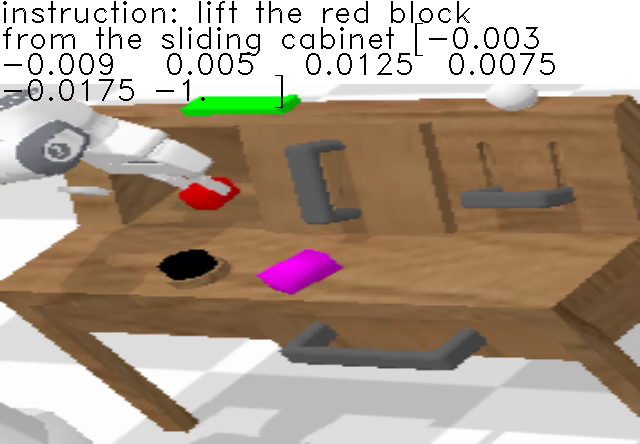} \\
\addlinespace[2.5pt]
\midrule
\addlinespace[4pt] Pull the handle to open the drawer. & \includegraphics[width=0.22\textwidth,valign=m]{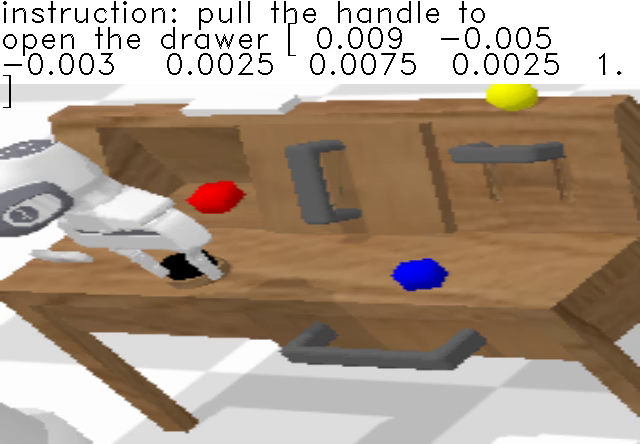} & 
$\rightarrow$ & 
\includegraphics[width=0.22\textwidth,valign=m]{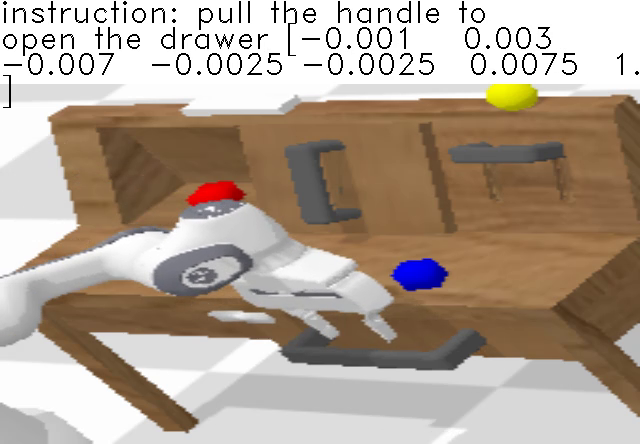} & 
$\rightarrow$ & 
\includegraphics[width=0.22\textwidth,valign=m]{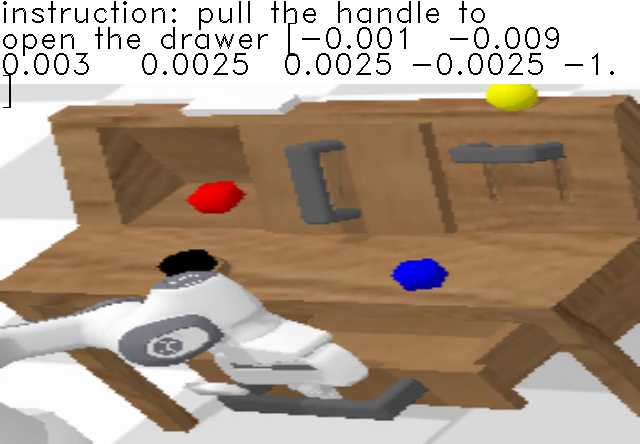} \\
\addlinespace[2.5pt]
\midrule 
%
\addlinespace[4pt] Push the sliding door to the left side. & \includegraphics[width=0.22\textwidth,valign=m]{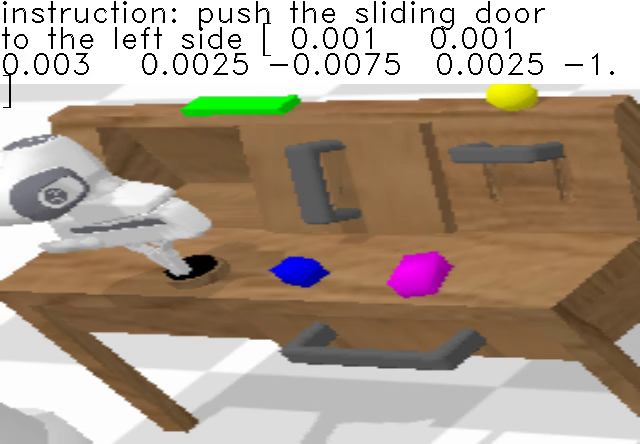} & 
$\rightarrow$ & 
\includegraphics[width=0.22\textwidth,valign=m]{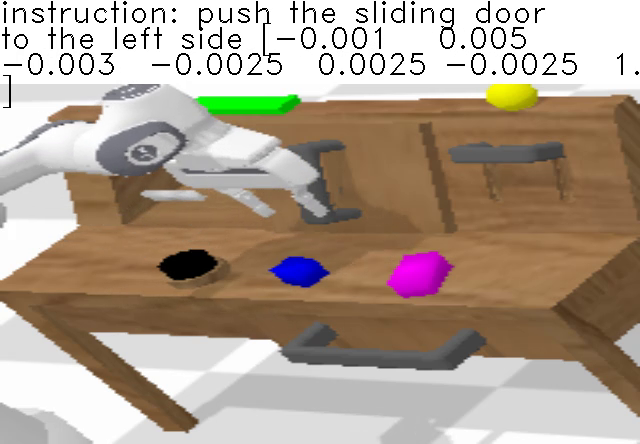} & 
$\rightarrow$ & 
\includegraphics[width=0.22\textwidth,valign=m]{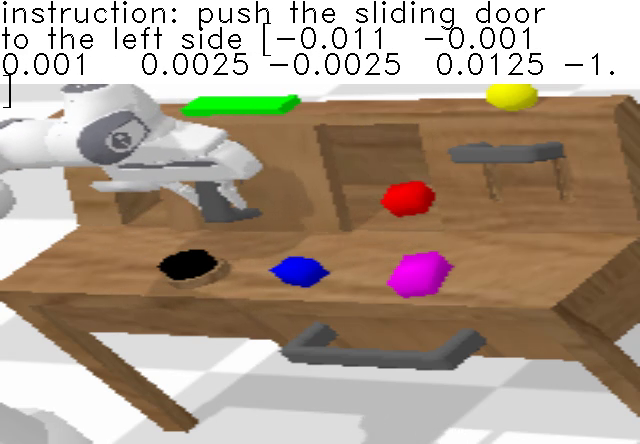} \\
\addlinespace[2.5pt]
\midrule
\addlinespace[4pt]
Push the sliding door to the right side. & 
\includegraphics[width=0.22\textwidth,valign=m]{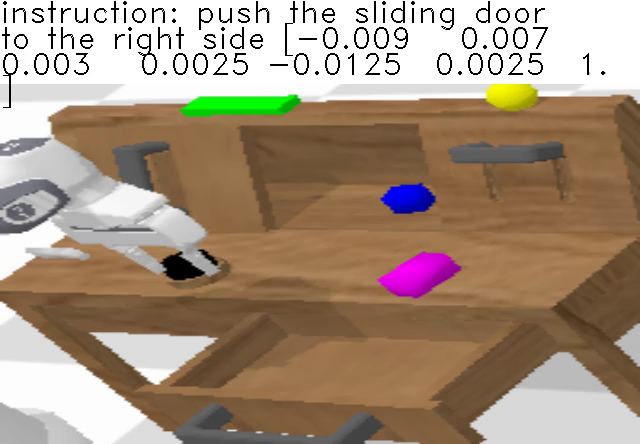} & 
$\rightarrow$ & 
\includegraphics[width=0.22\textwidth,valign=m]{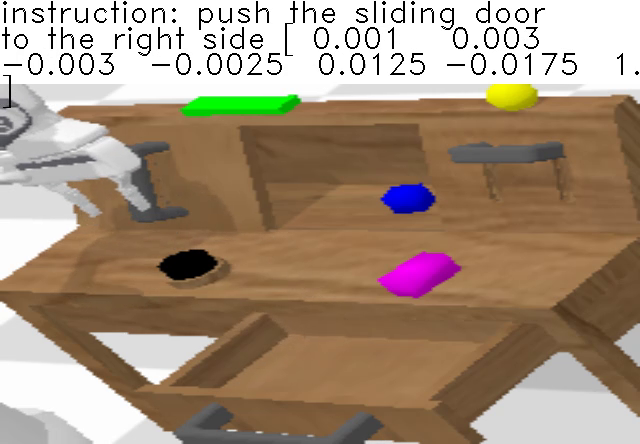} & 
$\rightarrow$ & 
\includegraphics[width=0.22\textwidth,valign=m]{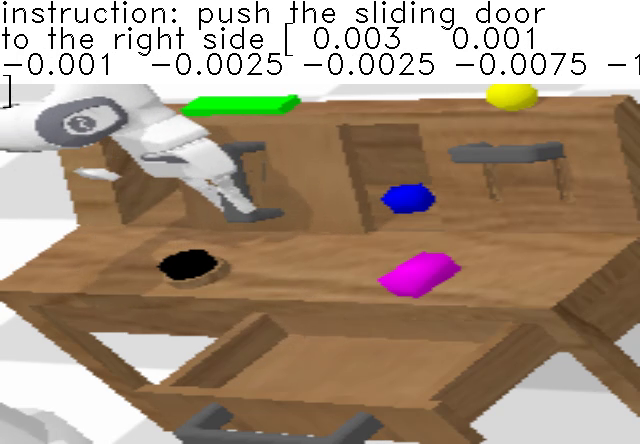} \\
\bottomrule
\end{tabular}
\caption{We show 5 unique demonstrations from CALVIN, where our model successfully follows the text instruction.  In addition to the high level instruction, we also show the low-level predicted actions of our agent above each frame.}
\label{tab:calvin_samples}

\end{table*}

\begin{table*}[ht]
\centering
\begin{tabular}{@{}m{3cm}ccm{4cm}@{}}
Task & Start frame & End frame & Model Output \\
\toprule
Video Captioning & 
\includegraphics[width=0.22\textwidth,valign=m]{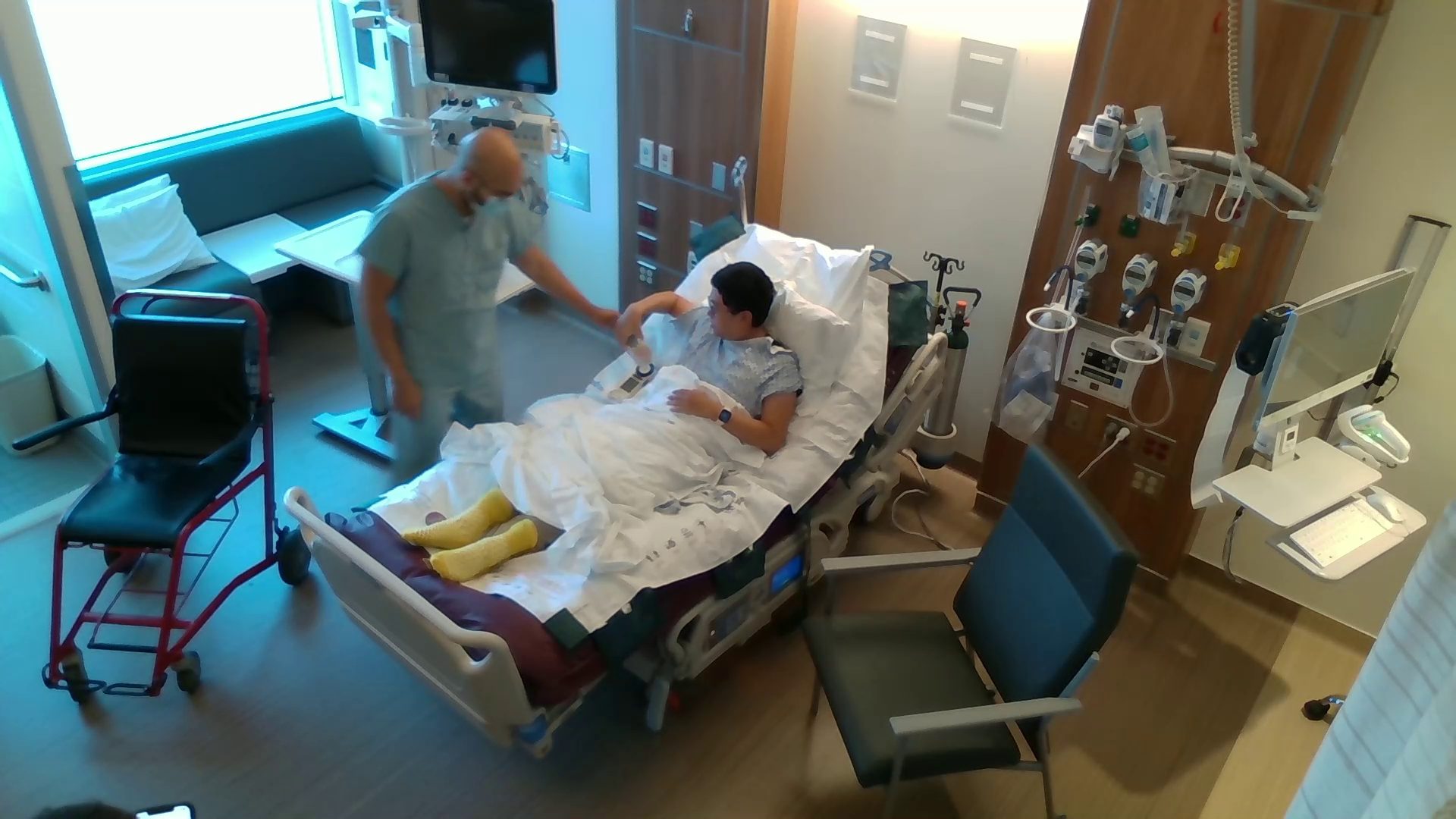} & \includegraphics[width=0.22\textwidth,valign=m]{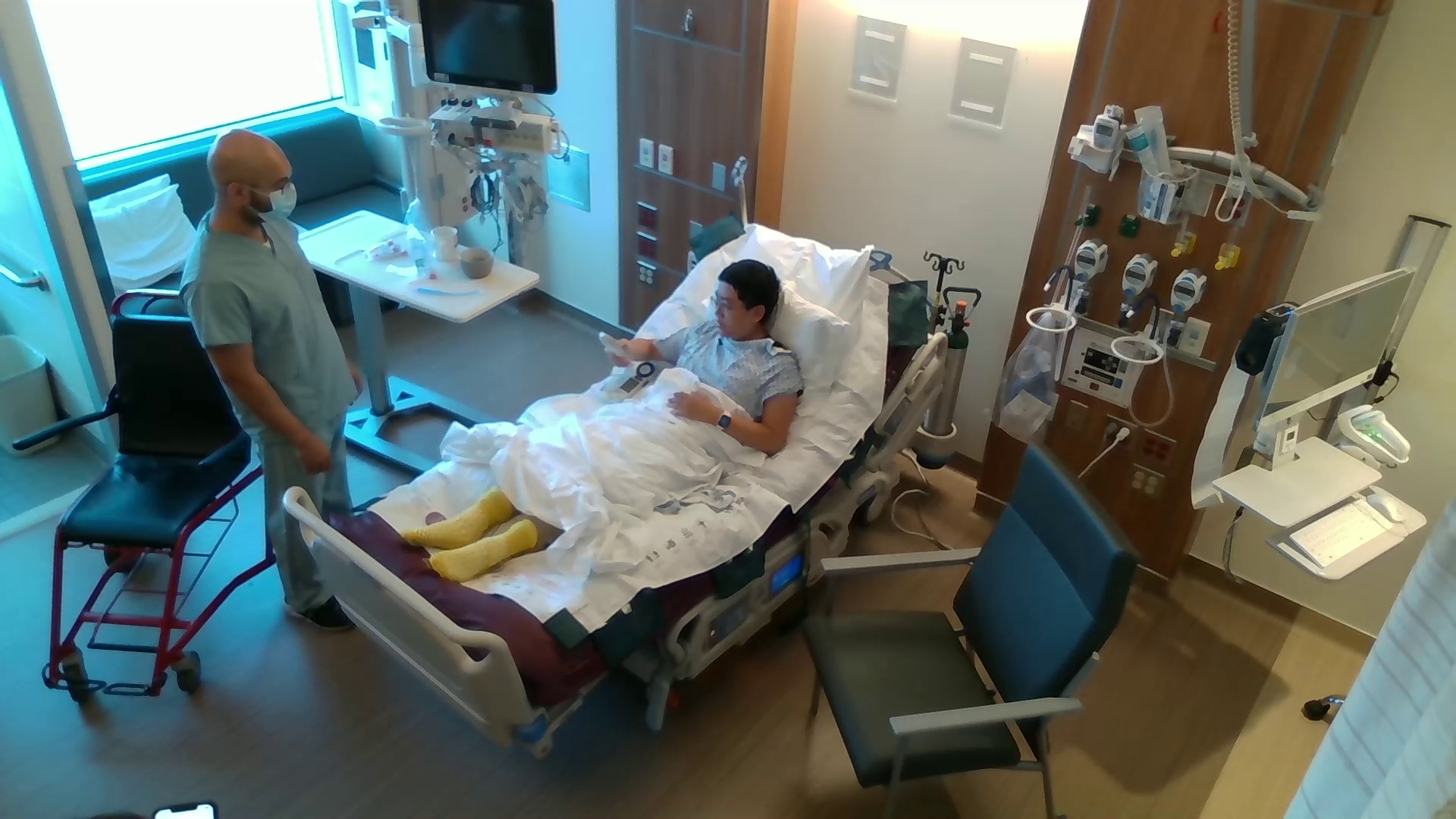} & The patient is awake and calm. The patient is cooperative. The patient is alert \\ 
\midrule

Video Question Answering & 
\includegraphics[width=0.22\textwidth,valign=m]{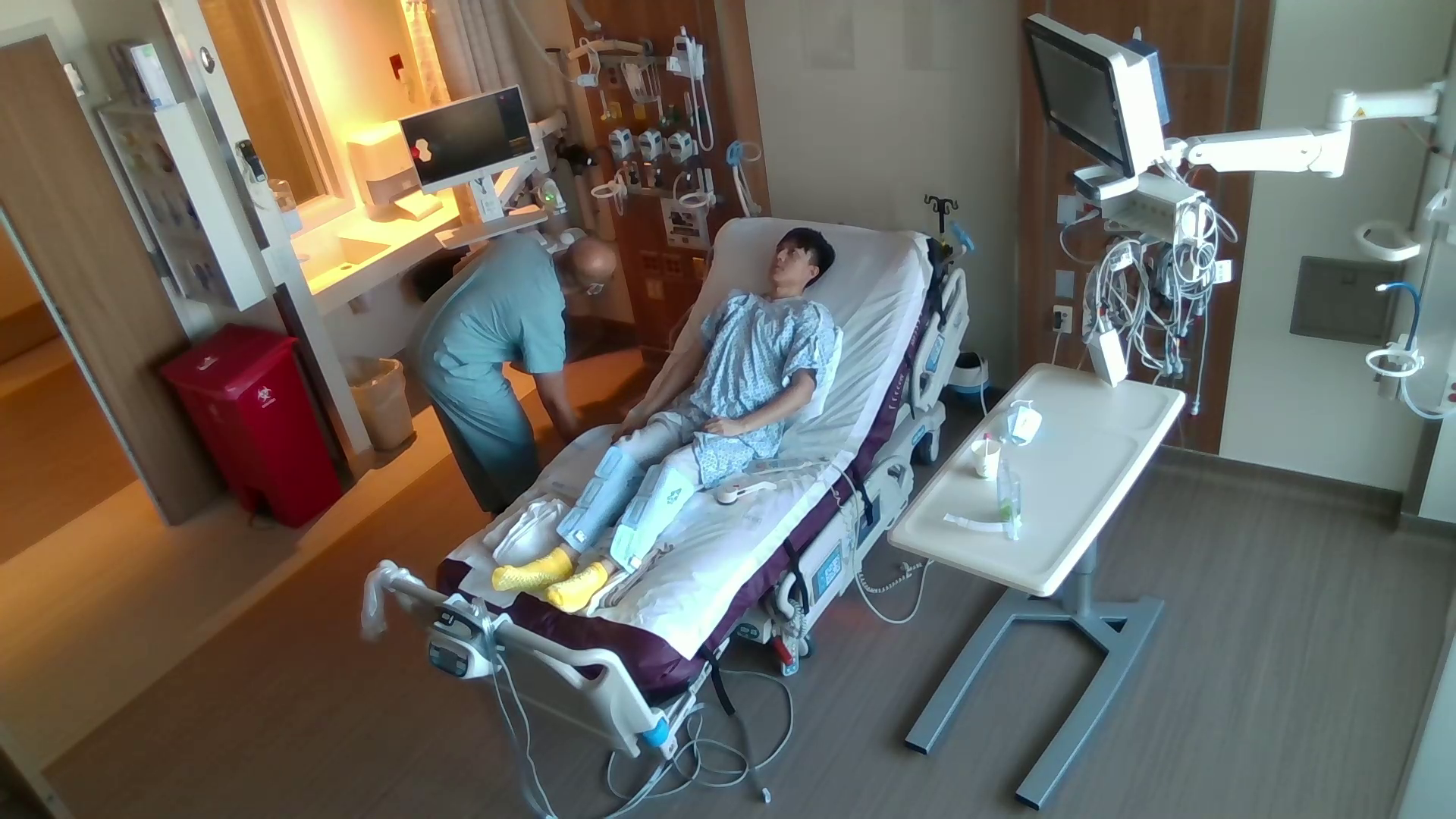} & \includegraphics[width=0.22\textwidth,valign=m]{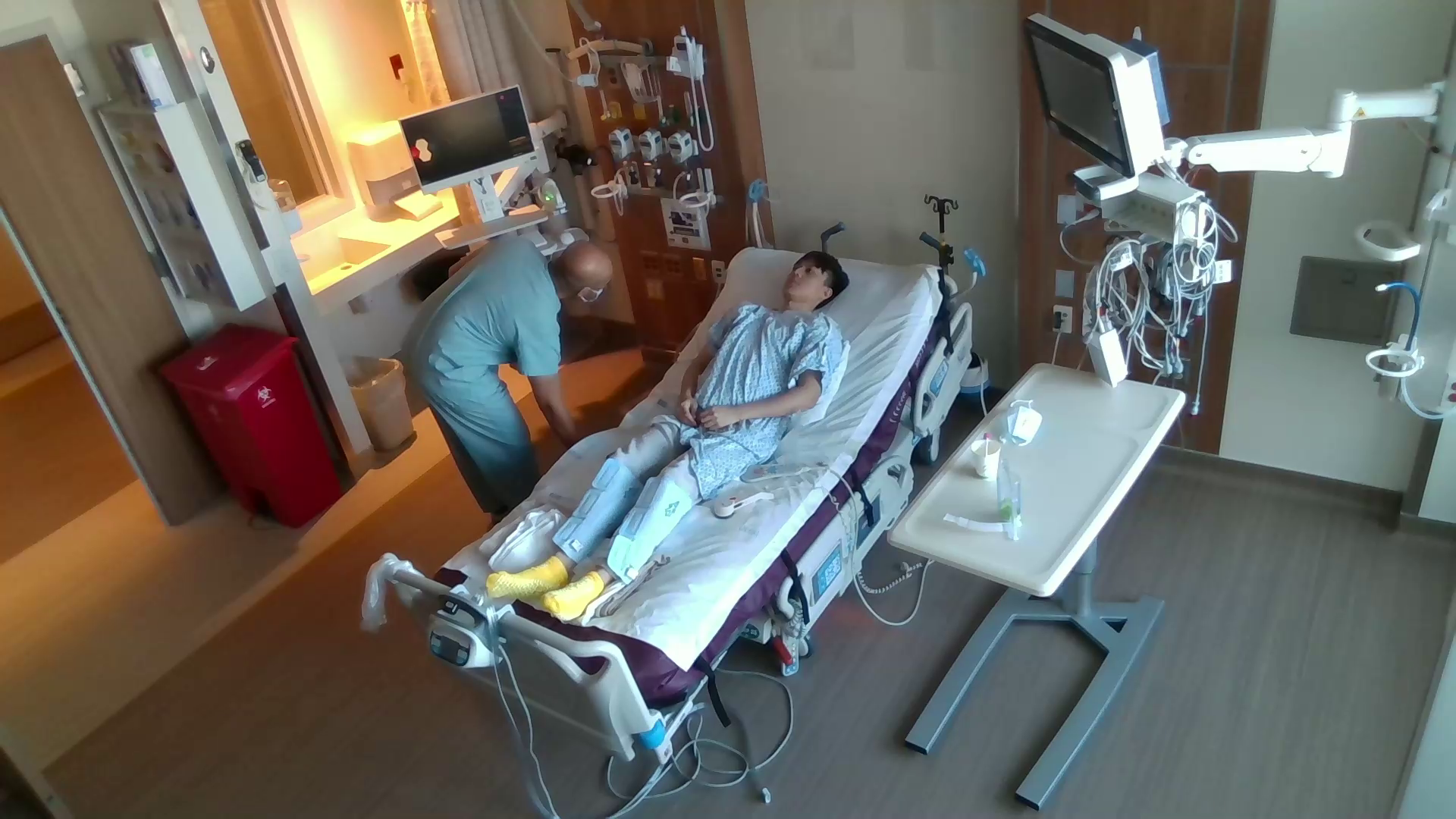} & Q: Where is the patient? A: patient is in deep sedation. The patient likely requires assistance. \\ 
\midrule

Action Recognition (RASS) & 
\includegraphics[width=0.22\textwidth,valign=m]{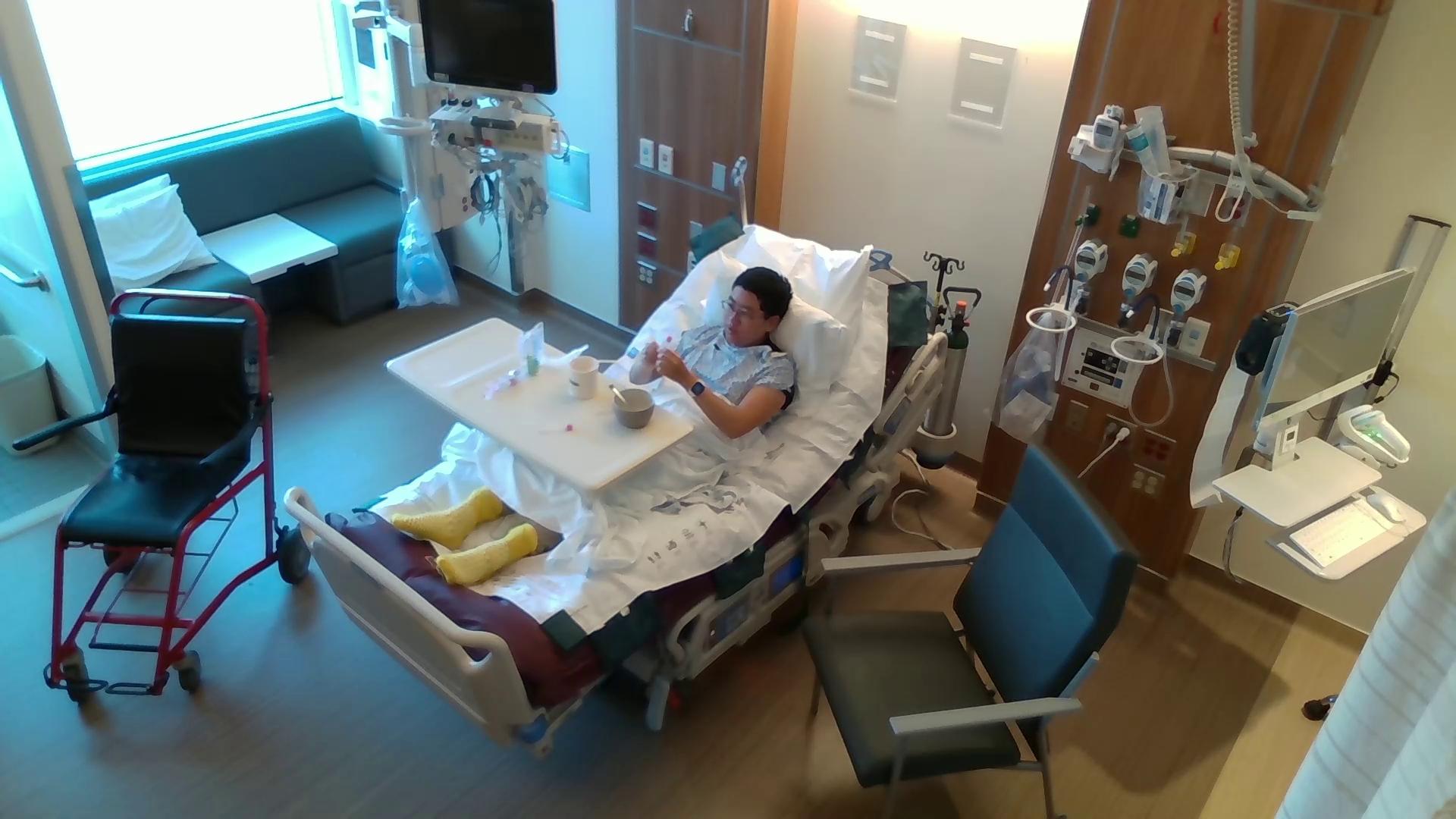} & \includegraphics[width=0.22\textwidth,valign=m]{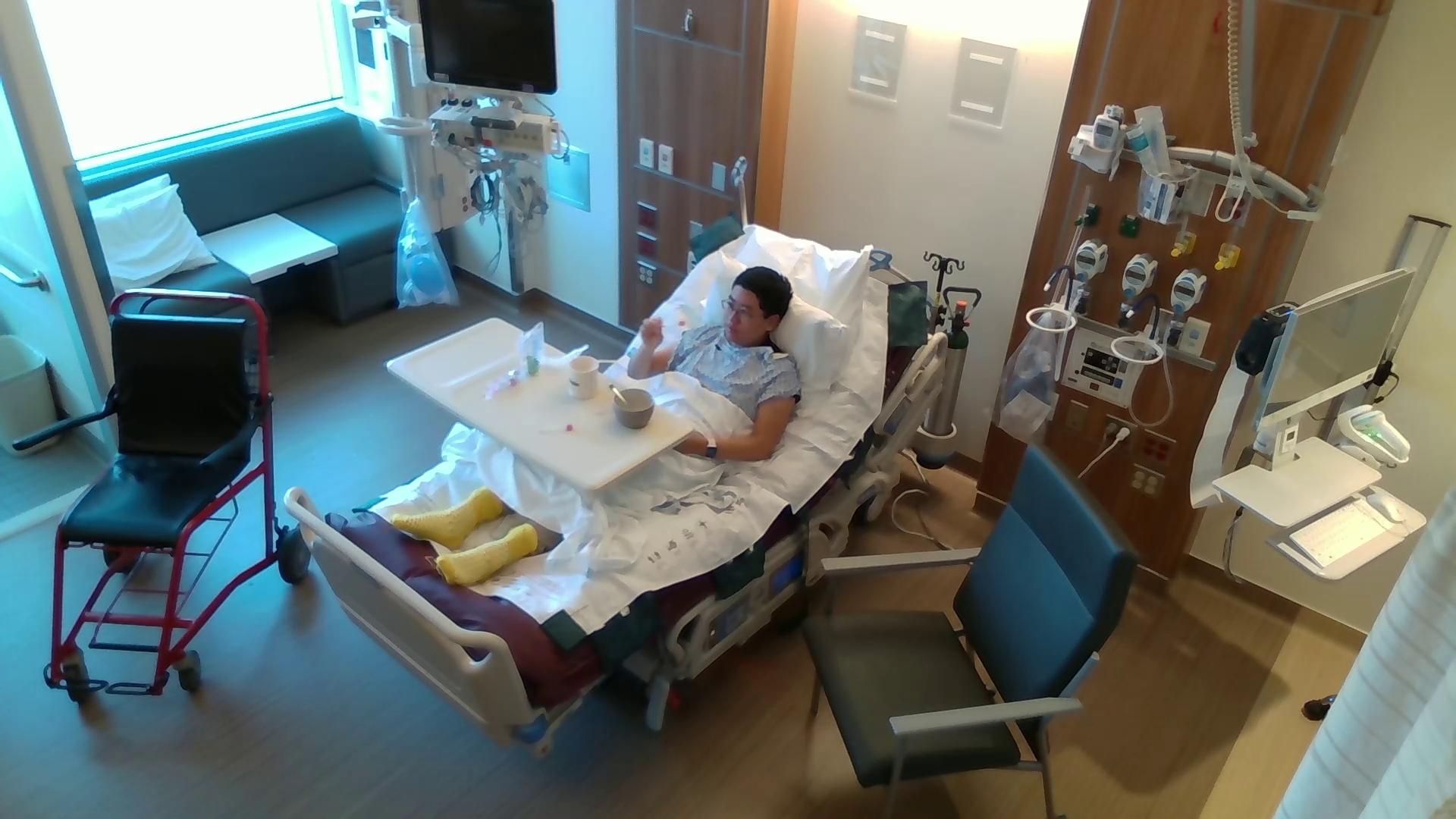} & 0 - Alert and calm \\ 
\midrule
Video Captioning & 
\includegraphics[width=0.22\textwidth,valign=m]{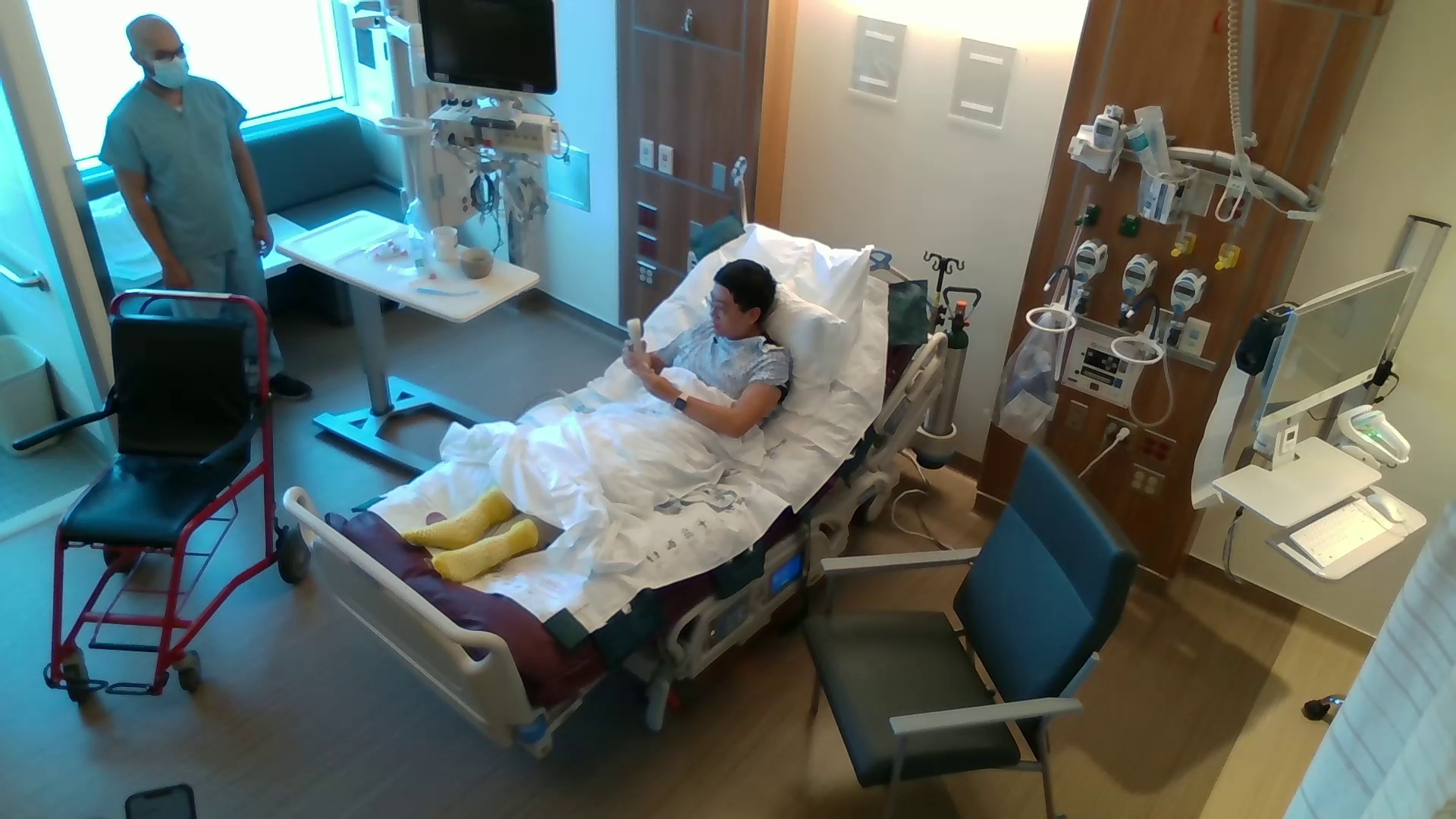} & \includegraphics[width=0.22\textwidth,valign=m]{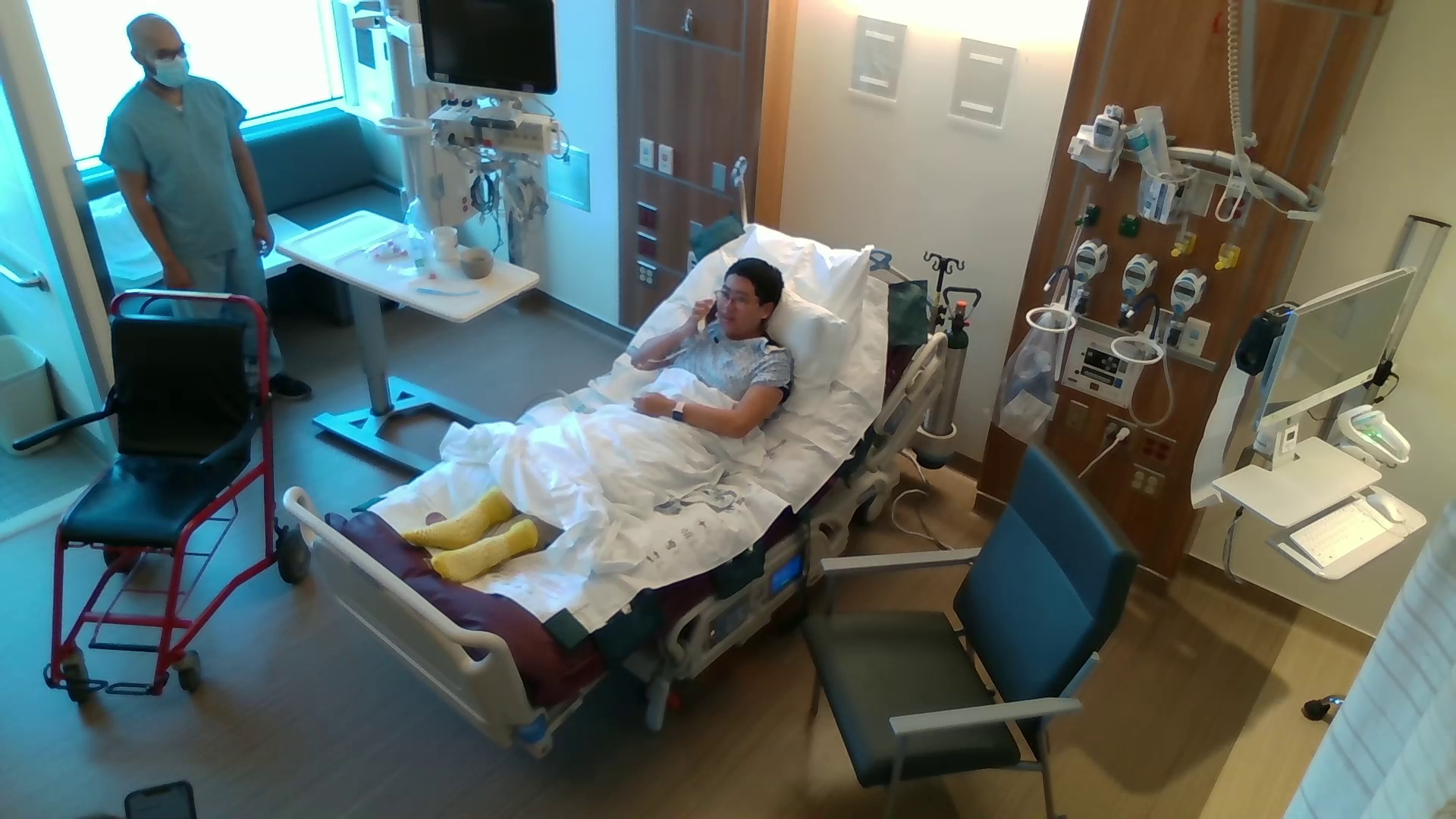} & The patient is awake and calm. They are speaking on the phone. \\ 

\bottomrule
\end{tabular}
\caption{We show 4 demonstrations of our agent model's outputs on a held-out Healthcare dataset that uses actors instead of actual patients. We demonstrate our model's outputs across 3 different tasks: video captioning, visual question answering, and RASS score prediction (action recognition). Due to the nature of our actor-collected example videos, the model predicts that the patient is awake and calm (RASS score of $0$)  for most video clips, despite only $~60\%$ of the training data containing RASS score of $0$.}
\label{tab:healthcare_samples}

\end{table*}

\begin{table*}[ht]
\centering
\begin{tabular}{@{}m{3cm}|c|@{}m{4.5cm}|@{}m{4.5cm}}
Text instruction & Start frame & Predicted Action &  Ground Truth Action \\
\toprule
the player is digging and placing dirt blocks to terraform the terrain around their house... & 
\includegraphics[width=0.22\textwidth,valign=m]{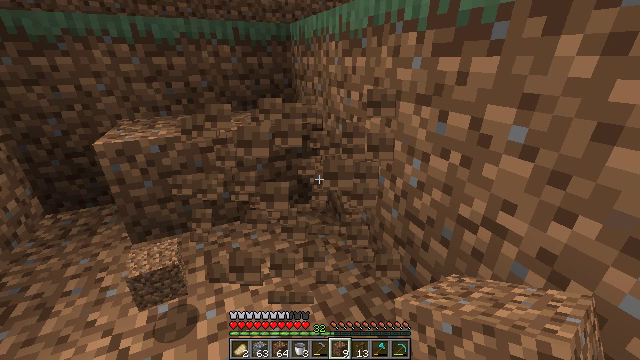} & 
[STARTACTION] [attack] [CAMERAX0] [CAMERAY-1] [ENDOFACTION] & 
[STARTACTION] [attack] [ENDOFACTION] \\
\addlinespace[2.5pt]
\midrule

the player is mining underground using a diamond pickaxe, gathering cobblestone, coal, iron ore... & 
\includegraphics[width=0.22\textwidth,valign=m]{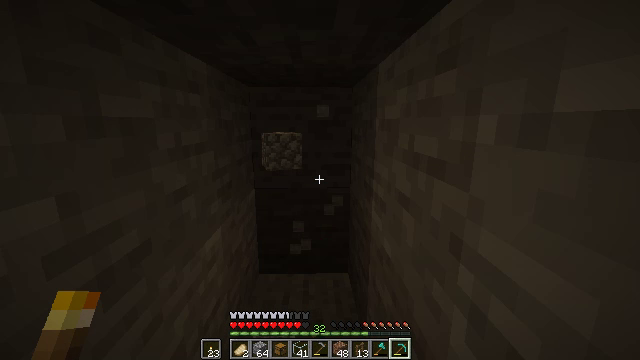} & 
[STARTACTION] [attack] [CAMERAX-3] [CAMERAY0] [ENDOFACTION] & 
[STARTACTION] [attack] [CAMERAX-3] [CAMERAY0] [ENDOFACTION] \\
\addlinespace[2.5pt]
\midrule

the minecraft player is moving around a village ... & 
\includegraphics[width=0.22\textwidth,valign=m]{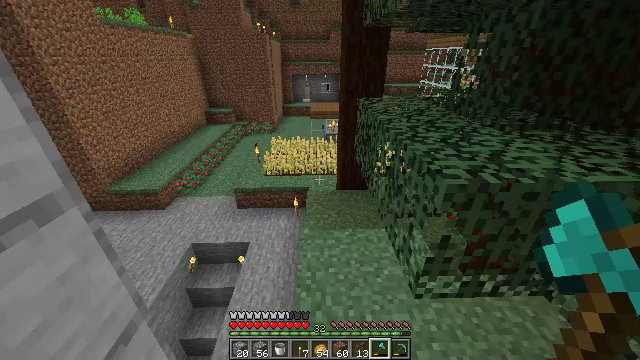} & 
[STARTACTION] [forward] [sprint] [ENDOFACTION] & 
[STARTACTION] [forward] [sprint] [ENDOFACTION] \\
\addlinespace[2.5pt]
\midrule

the player is using a brewing stand ... & 
\includegraphics[width=0.22\textwidth,valign=m]{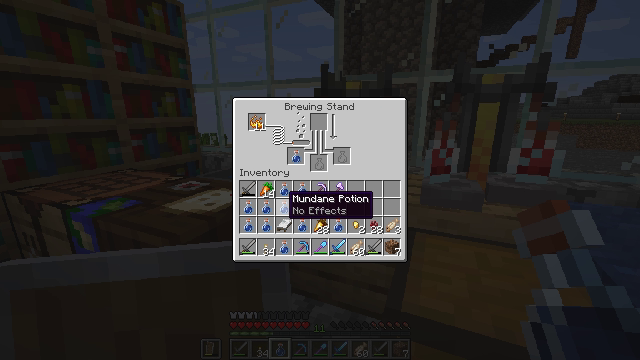} & 
[STARTACTION] [sneak] [use] [ENDOFACTION] & 
[STARTACTION] [sneak] [ENDOFACTION] \\
\addlinespace[2.5pt]
\midrule

the player is ... terraforming by digging ... & 
\includegraphics[width=0.22\textwidth,valign=m]{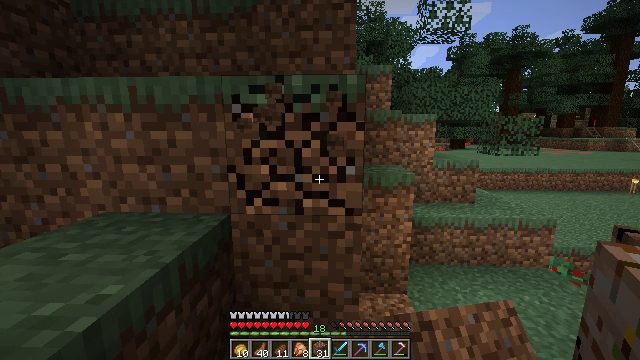} & 
[STARTACTION] [attack] [ENDOFACTION] & 
[STARTACTION] [attack] [ENDOFACTION] \\
\addlinespace[2.5pt]

\bottomrule
\end{tabular}
\caption{We show 5 demonstrations from a held-out Minecraft dataset. In addition to the high level instruction, we show the low-level predicted actions and ground truth actions. We truncate the instructions to show only the parts relevant to the current frames. The most common errors are slight differences in camera movements and occasionally performing unnecessary actions. Note that sometimes the ground truth values are not the only valid actions; for instance, the fourth example predicts that the player will click the bottle, which happens a few frames later in the ground truth trajectory.}
\label{tab:minecraft_samples}

\end{table*}

\begin{table*}[ht]
\centering
\begin{tabular}{@{}m{3cm}|c|@{}m{4.5cm}|@{}m{4.5cm}}
Text instruction & Start frame & Predicted Action &  Ground Truth Action \\
\toprule
the player is using a character with a sword to fight enemies and collect power cells ... & 
\includegraphics[width=0.22\textwidth,valign=m]{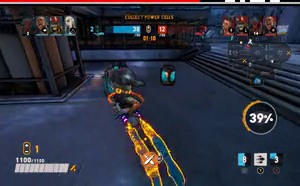} & 
[STARTACTION] [lockon][meleeattack] [lrot214] [lmag4] [ENDOFACTION] & 
[STARTACTION] [lockon][meleeattack] [lrot213] [lmag4] [ENDOFACTION] \\
\addlinespace[2.5pt]
\midrule

the player is riding a hoverboard-like vehicle ...  avoiding or attacking enemy players ... & 
\includegraphics[width=0.22\textwidth,valign=m]{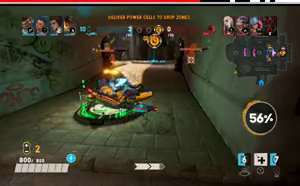} & 
[STARTACTION] [lockon][meleeattack] [lrot204] [lmag4] [ENDOFACTION] & 
[STARTACTION] [lockon][meleeattack] [lrot201] [lmag4] [ENDOFACTION] \\
\addlinespace[2.5pt]
\midrule

the player starts by descending some stairs towards an open area where they engage in combat with an enemy player ... & 
\includegraphics[width=0.22\textwidth,valign=m]{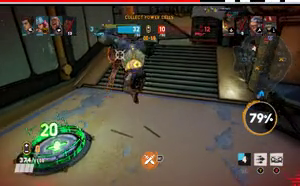} & 
[STARTACTION] [jump] [lockon][specialability1] [lrot199] [lmag4] [ENDOFACTION] & 
[STARTACTION] [jump] [lockon][meleeattack] [lrot201] [lmag4] [ENDOFACTION] \\
\addlinespace[2.5pt]
\midrule

the player ... captures an objective point while fighting off multiple opponents ... & 
\includegraphics[width=0.22\textwidth,valign=m]{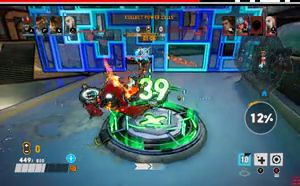} & 
[STARTACTION] [lockon][meleeattack] [lrot63] [lmag4] [ENDOFACTION] & 
[STARTACTION] [lockon][meleeattack] [lrot63] [lmag4] [ENDOFACTION] \\
\addlinespace[2.5pt]
\midrule

a bleeding edge player is controlling a robot character with a sword ... engaging in combat with enemy players ... & 
\includegraphics[width=0.22\textwidth,valign=m]{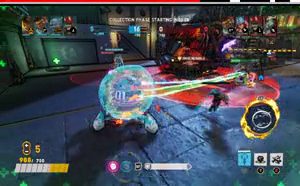} & 
[STARTACTION] [evade] [lrot236] [lmag4] [ENDOFACTION] & 
[STARTACTION] [evade] [lrot236] [lmag4] [ENDOFACTION] \\
\addlinespace[2.5pt]

\bottomrule
\end{tabular}
\caption{We show 5 unique demonstrations from a held-out Bleeding Edge dataset. In addition to the high level instruction, we show the low-level predicted actions and ground truth actions. We truncate the instructions to show only the parts relevant to the current frames. The most common errors are slight deviations from the precise value of the joysticks, which are naturally noisy. Some other errors include predicting the wrong type of attack, though this typically happens in situations where multiple attacks are still valid.}
\label{tab:be_samples}

\end{table*}

\end{document}